\documentclass[10pt,twocolumn,letterpaper]{article}

\usepackage{cvpr}
\usepackage{times}
\usepackage{epsfig}
\usepackage{graphicx}
\usepackage{epstopdf}
\usepackage{amsmath}
\usepackage{amssymb}
\usepackage{booktabs}
\usepackage{slashbox}
\usepackage{authblk}
\usepackage[font=small, labelfont=bf,up, textfont=up]{caption}

\newcommand*\samethanks[1][\value{footnote}]{\footnotemark[#1]}

\makeatletter
\renewcommand\AB@affilsepx{  \hspace{5 mm}  \protect\Affilfont}
\makeatother

\cvprfinalcopy 

\def\cvprPaperID{2015} 

\ifcvprfinal\fi
\begin{document}

\title{BigHand2.2M Benchmark: Hand Pose Dataset and State of the Art Analysis}

\author[]{Shanxin Yuan$^1$\thanks{indicates equal contribution} $\qquad$ Qi Ye$^1$\samethanks{} $\qquad$ Bj{\"o}rn Stenger$^2$ $\qquad$ Siddhant Jain$^3$ $\qquad$ Tae-Kyun Kim}

\affil[1]{Imperial College London} 
\affil[2]{Rakuten Institute of Technology} 
\affil[3]{IIT Jodhpur}

\maketitle
\thispagestyle{empty}

\begin{abstract}

In this paper we introduce a large-scale hand pose dataset, collected using a novel capture method.
Existing datasets are either generated synthetically or captured using depth sensors: synthetic datasets exhibit a certain level of appearance difference from real depth images, and real datasets are limited in quantity and coverage, mainly due to the difficulty to annotate them. 
We propose a tracking system with six 6D magnetic sensors and inverse kinematics to automatically obtain 21-joints hand pose annotations of depth maps captured with minimal restriction on the range of motion.
The capture protocol aims to fully cover the natural hand pose space. As shown in embedding plots, the new dataset exhibits a significantly wider and denser range of hand poses compared to existing benchmarks. Current state-of-the-art methods are evaluated on the dataset, and we demonstrate significant improvements in cross-benchmark performance. We also show significant improvements in egocentric hand pose estimation with a CNN trained on the new dataset.

\end{abstract}

\vspace{-5mm}
\section{Introduction} 
\label{paraIntro}
\vspace{-1mm}

\begin{figure*}[ht]
	\centering
	
		\includegraphics[trim=2cm 1cm 2cm 1cm, clip=true,width=0.09\textwidth]{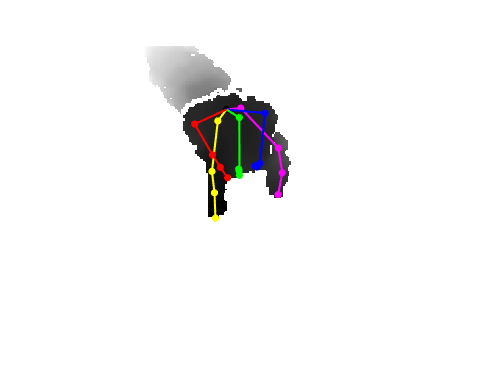}
		\includegraphics[trim=2cm 1cm 2cm 1cm, clip=true,width=0.09\textwidth]{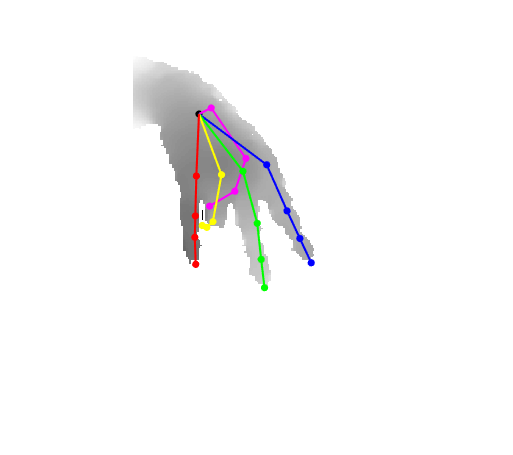}
		\includegraphics[trim=2cm 1cm 2cm 1cm, clip=true,width=0.09\textwidth]{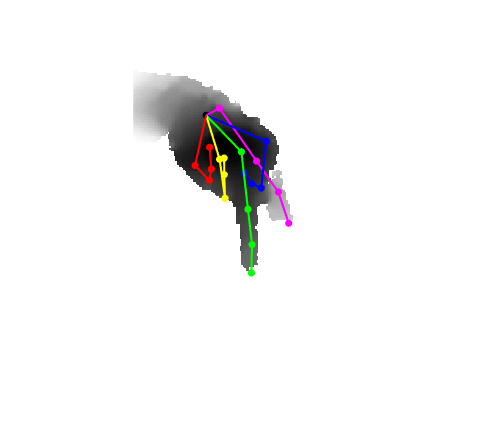}
		\includegraphics[trim=2cm 1cm 2cm 1cm, clip=true,width=0.09\textwidth]{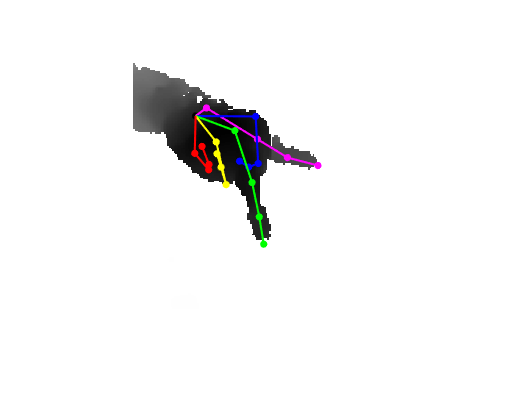}
		\includegraphics[trim=2cm 1cm 2cm 1cm, clip=true,width=0.09\textwidth]{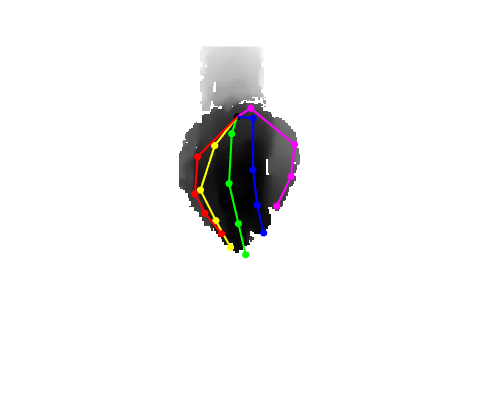}
		\includegraphics[trim=2cm 1cm 2cm 1cm, clip=true,width=0.09\textwidth]{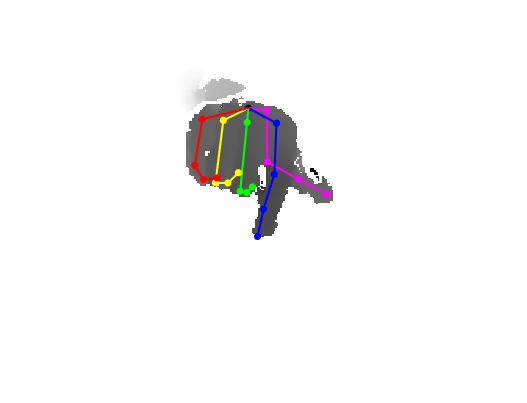}
		\includegraphics[trim=2cm 1cm 2cm 1cm, clip=true,width=0.09\textwidth]{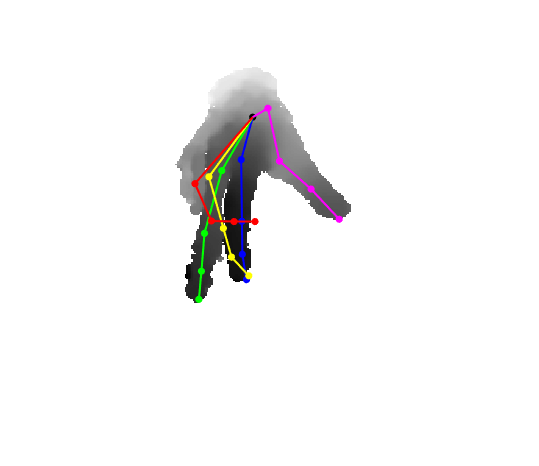}
		\includegraphics[trim=2cm 1cm 2cm 1cm, clip=true,width=0.09\textwidth]{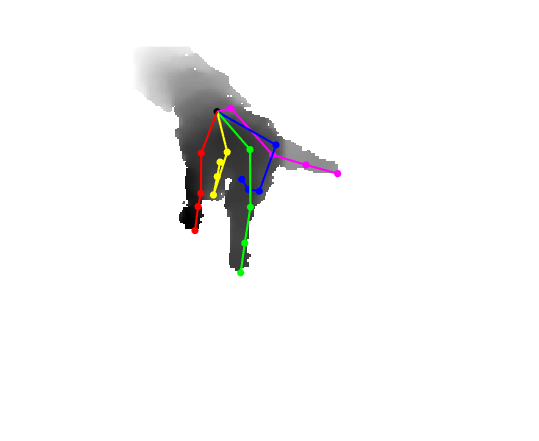}
		\includegraphics[trim=2cm 1cm 2cm 1cm, clip=true,width=0.09\textwidth]{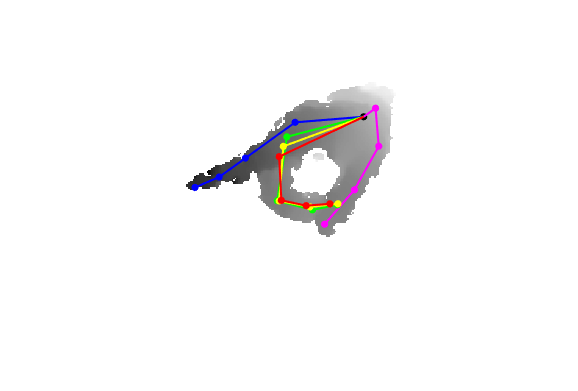}
		\includegraphics[trim=2cm 1cm 2cm 1cm, clip=true,width=0.09\textwidth]{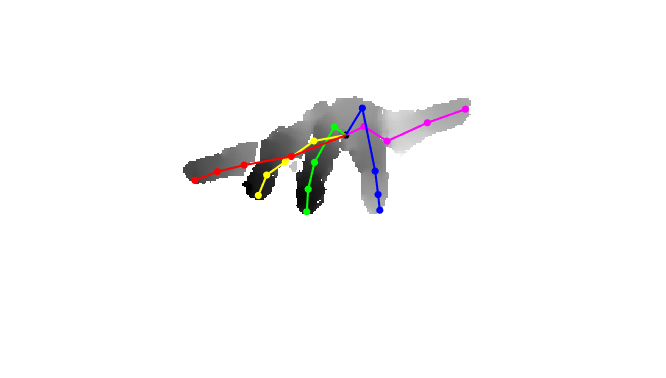}\hfill \\	

		\includegraphics[trim=2cm 1cm 2cm 1cm, clip=true,width=0.09\textwidth]{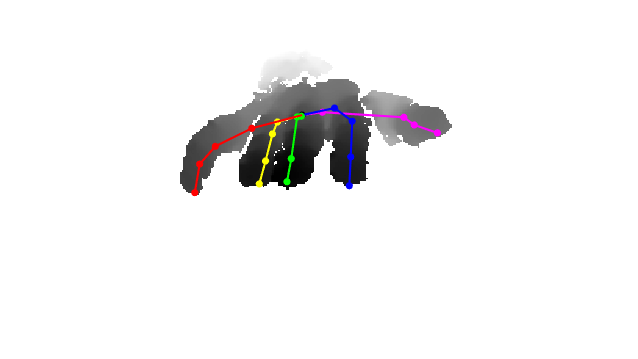}
		\includegraphics[trim=2cm 1cm 2cm 1cm, clip=true,width=0.09\textwidth]{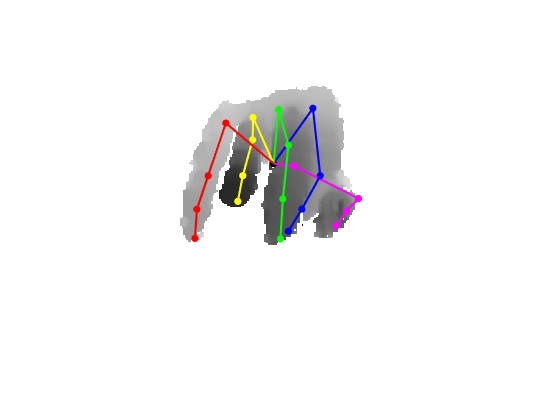}
		\includegraphics[trim=2cm 1cm 2cm 1cm, clip=true,width=0.09\textwidth]{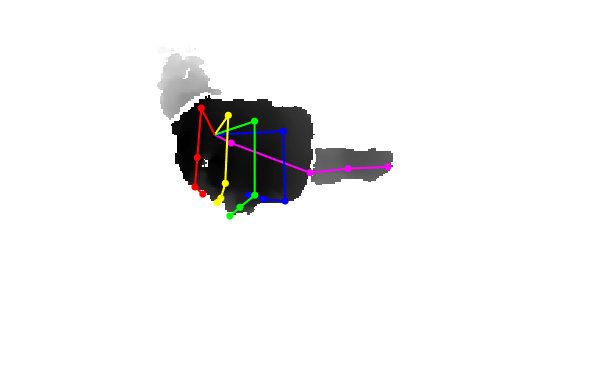}
		\includegraphics[trim=2cm 1cm 2cm 1cm, clip=true,width=0.09\textwidth]{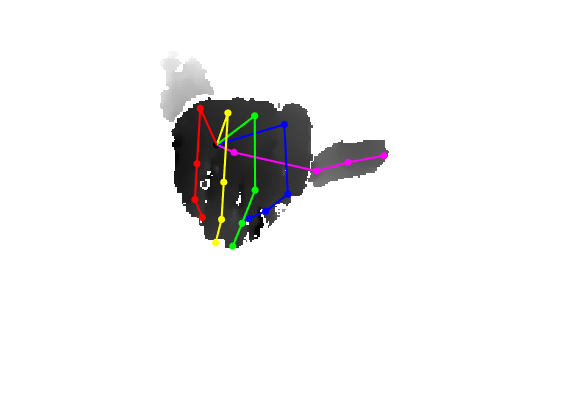}
		\includegraphics[trim=2cm 1cm 2cm 1cm, clip=true,width=0.09\textwidth]{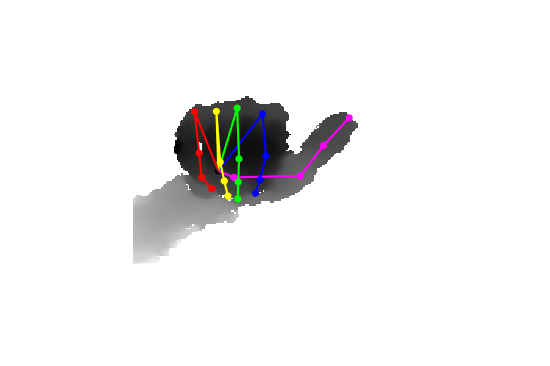}
		\includegraphics[trim=2cm 1cm 2cm 1cm, clip=true,width=0.09\textwidth]{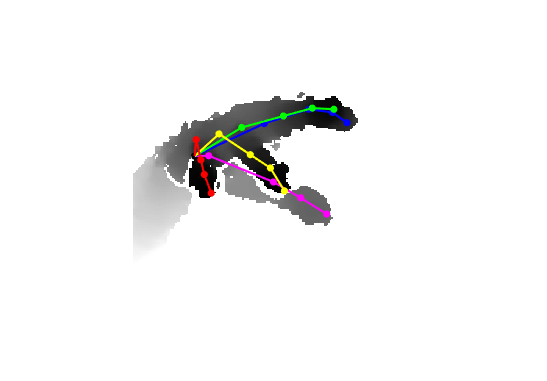}
		\includegraphics[trim=2cm 1cm 2cm 1cm, clip=true,width=0.09\textwidth]{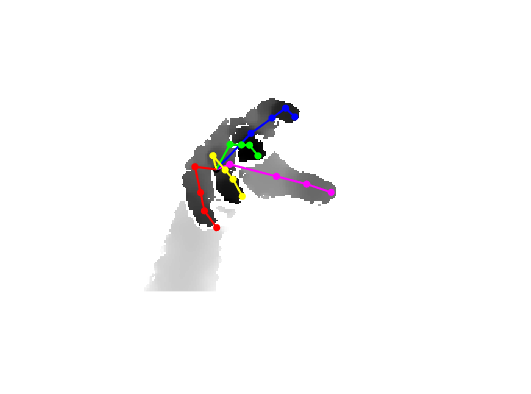}
		\includegraphics[trim=2cm 1cm 2cm 1cm, clip=true,width=0.09\textwidth]{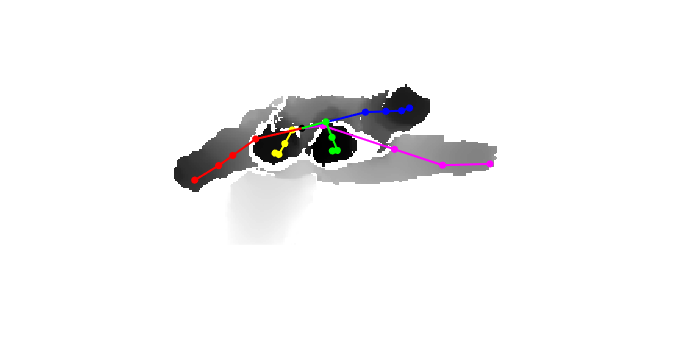}
		\includegraphics[trim=2cm 1cm 2cm 1cm, clip=true,width=0.09\textwidth]{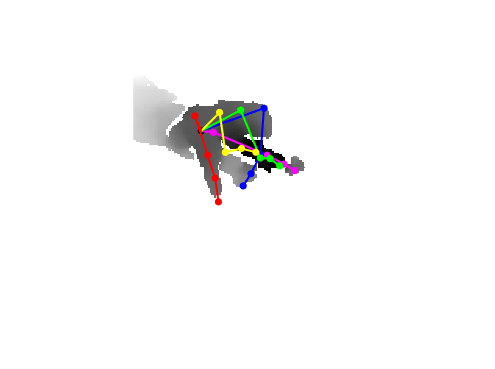}
		\includegraphics[trim=2cm 1cm 2cm 1cm, clip=true,width=0.09\textwidth]{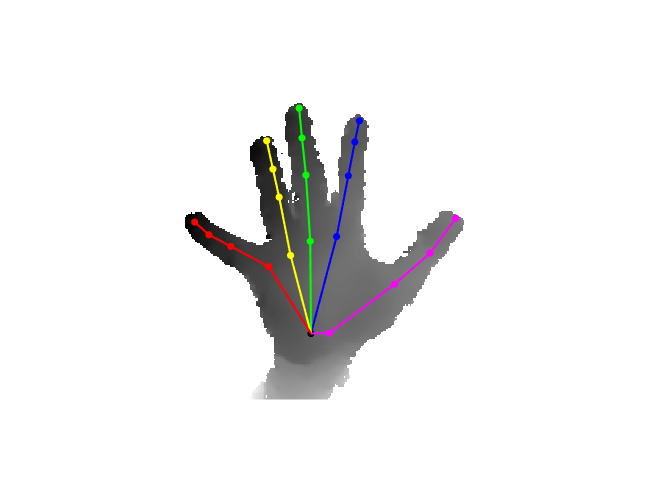}\hfill \\	
		\includegraphics[trim=2cm 1cm 2cm 1cm, clip=true,width=0.09\textwidth]{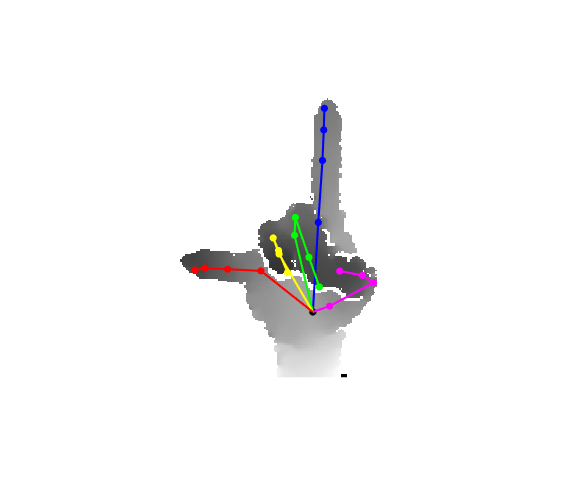}
		\includegraphics[trim=2cm 1cm 2cm 1cm, clip=true,width=0.09\textwidth]{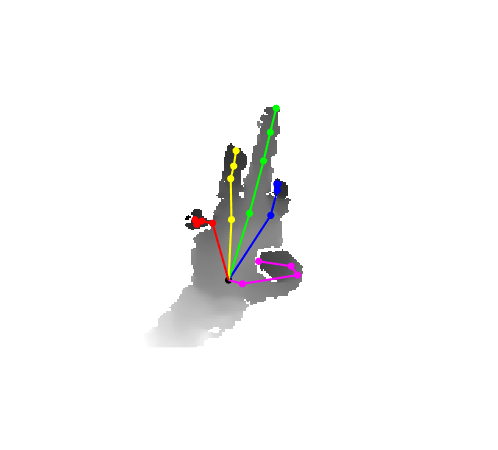}
		\includegraphics[trim=2cm 1cm 2cm 1cm, clip=true,width=0.09\textwidth]{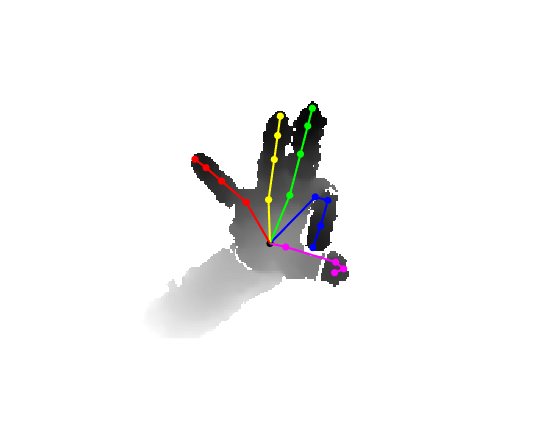}
		\includegraphics[trim=2cm 1cm 2cm 1cm, clip=true,width=0.09\textwidth]{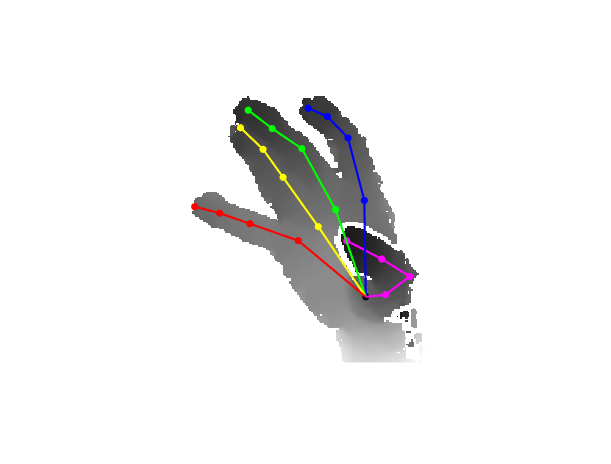}
		\includegraphics[trim=2cm 1cm 2cm 1cm, clip=true,width=0.09\textwidth]{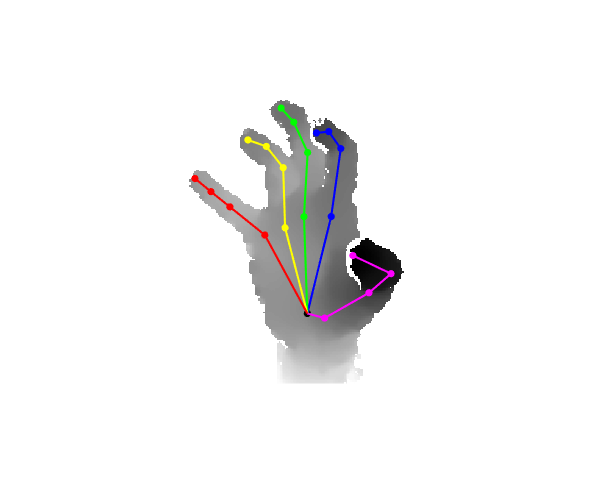}
		\includegraphics[trim=2cm 1cm 2cm 1cm, clip=true,width=0.09\textwidth]{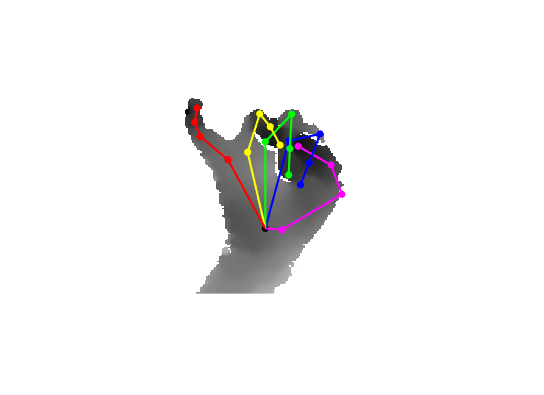}
		\includegraphics[trim=2cm 1cm 2cm 1cm, clip=true,width=0.09\textwidth]{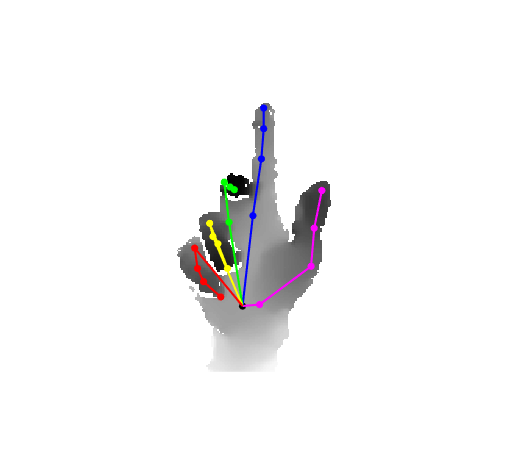}
		\includegraphics[trim=2cm 1cm 2cm 1cm, clip=true,width=0.09\textwidth]{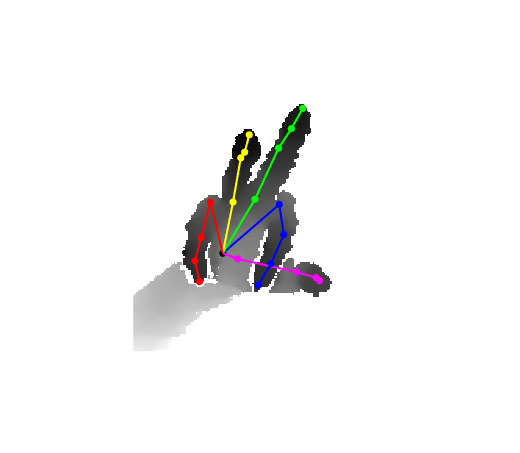}
		\includegraphics[trim=2cm 1cm 2cm 1cm, clip=true,width=0.09\textwidth]{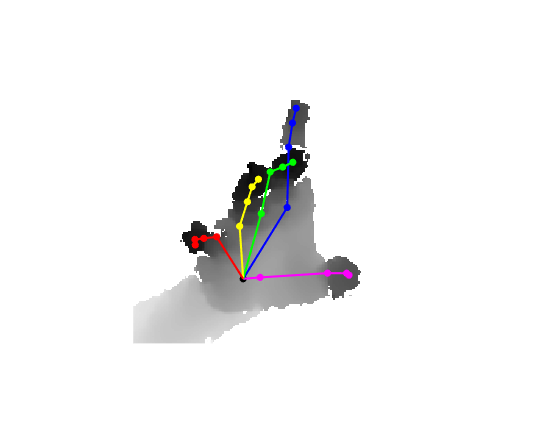}
		\includegraphics[trim=2cm 1cm 2cm 1cm, clip=true,width=0.09\textwidth]{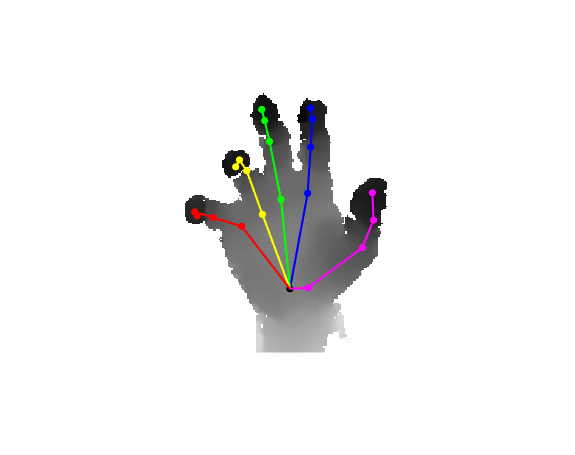}\hfill \\	
		
		\includegraphics[trim=1.5cm 0.5cm 1.5cm 0.5cm, clip=true,width=0.09\textwidth]{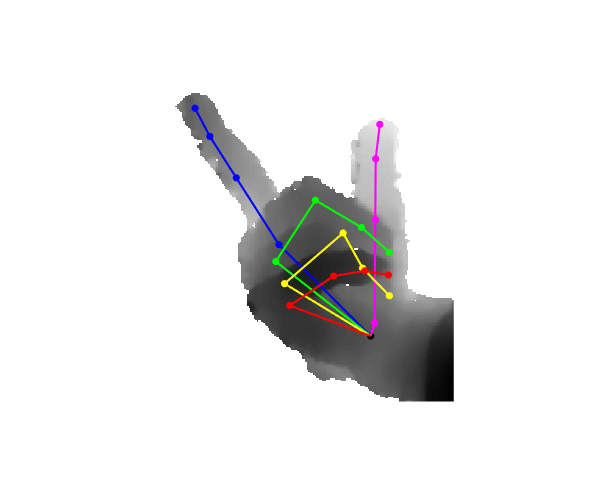}
		\includegraphics[trim=1.5cm 0.5cm 1.5cm 0.5cm, clip=true,width=0.09\textwidth]{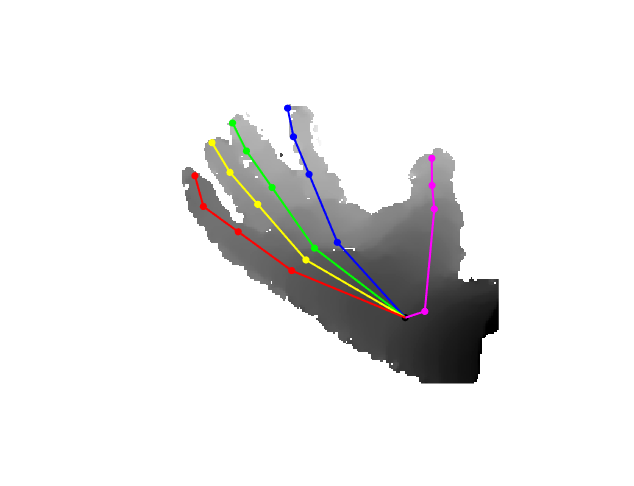}
		\includegraphics[trim=1.5cm 0.5cm 1.5cm 0.5cm, clip=true,width=0.09\textwidth]{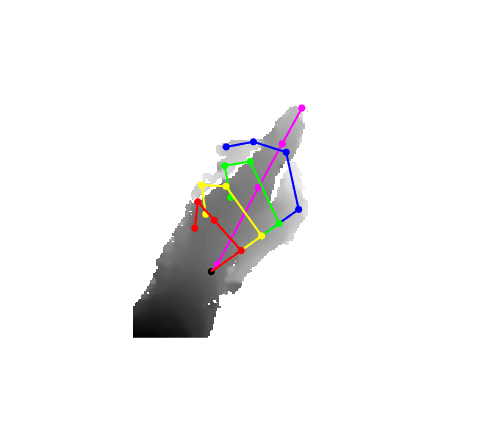}		
		\includegraphics[trim=1.5cm 0.5cm 1.5cm 0.5cm, clip=true,width=0.09\textwidth]{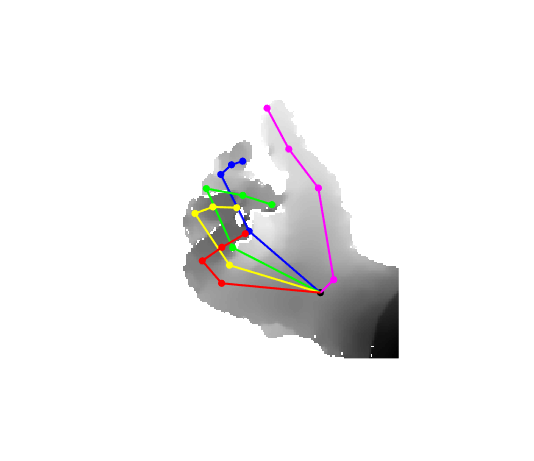}
		\includegraphics[trim=1.5cm 0.5cm 1.5cm 0.5cm, clip=true,width=0.09\textwidth]{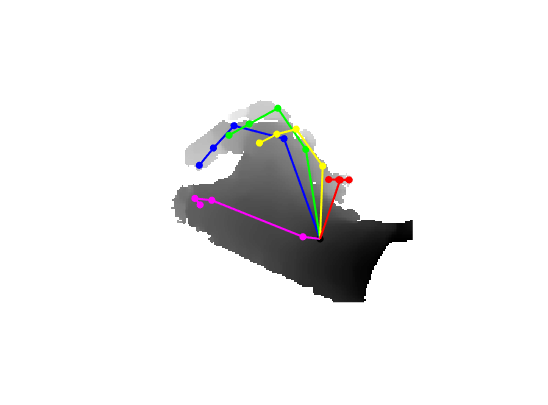}
		\includegraphics[trim=1.5cm 0.5cm 1.5cm 0.5cm, clip=true,width=0.09\textwidth]{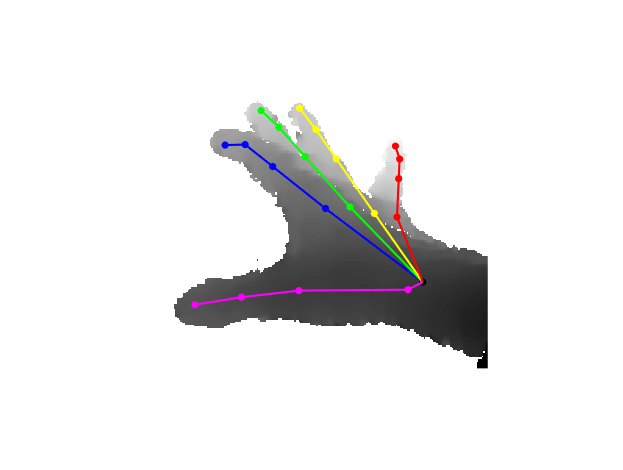}
		\includegraphics[trim=1.5cm 0.5cm 1.5cm 0.5cm, clip=true,width=0.09\textwidth]{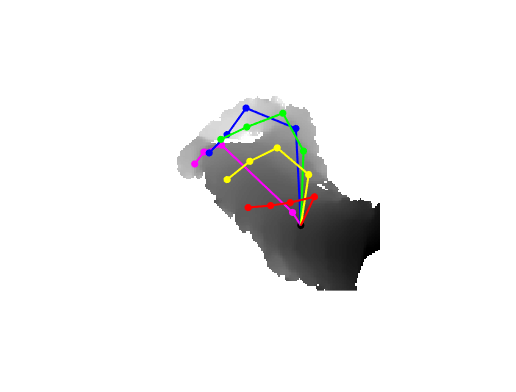}
		\includegraphics[trim=1.5cm 0.5cm 1.5cm 0.5cm, clip=true,width=0.09\textwidth]{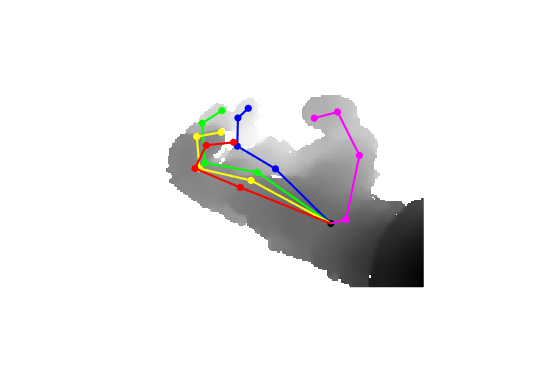}
		\includegraphics[trim=1.5cm 0.5cm 1.5cm 0.5cm, clip=true,width=0.09\textwidth]{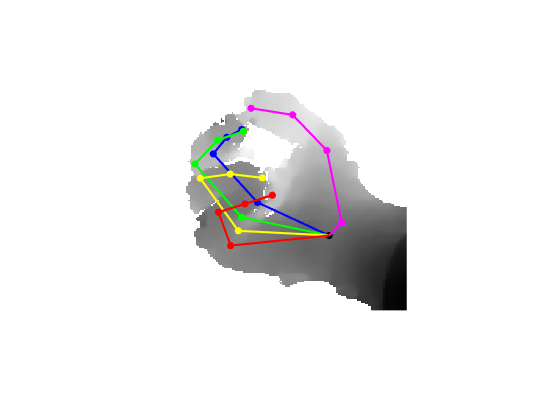}
		\includegraphics[trim=1.5cm 0.5cm 1.5cm 0.5cm, clip=true,width=0.09\textwidth]{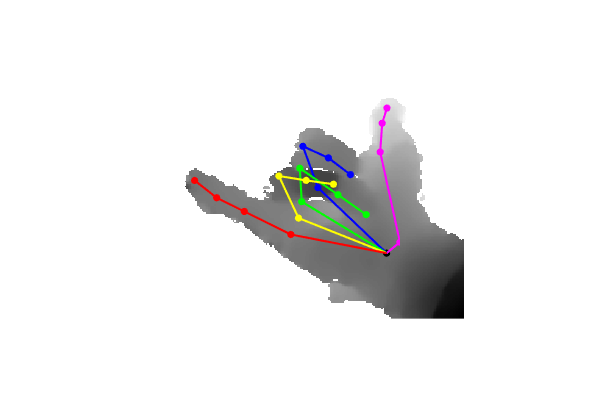}\hfill \\		
		
	 \caption{{\bf Example images from the {\it BigHand2.2M} dataset}. The dataset covers the range of hand poses that can be assumed without applying external forces to the hand. The accuracy of joint annotations is higher than in previous benchmark datasets.}
     \label{pic:challengingposes}
\end{figure*}

There has been significant progress in the area of hand pose estimation in the recent past and a number of systems have been proposed~\cite{keskin2012hand,liang2014parsing,neverova2015hand,oberweger2015training,oikonomidis2011efficient,qian2014realtime,sharp2015accurate,tang2013real,li20153d,choi2015collaborative,jang20153d,wan2016hand,khamis2015learning}. However, as noted in \cite{oberweger2016efficiently}, existing benchmarks \cite{qian2014realtime,sun2015cascade,tang2014latent,tompson2014real,wetzler2015rule} are restricted in terms of number of annotated images, annotation accuracy, articulation coverage, and variation in hand shape and viewpoint. 

The current state of the art for hand pose estimation employs deep neural networks to estimate hand pose from input data~\cite{tompson2014real,oberweger2015hands,oberweger2015training,zhou2016model,ge2016robust,ye2016spatial}. It has been shown that these methods scale well with the size of the training dataset. The availability of a large-scale, accurately annotated dataset is therefore a key factor for advancing the field. Manual annotation has been the bottleneck for creating large-scale benchmarks~\cite{qian2014realtime,handtracker_iccv2013}. This method is not only labor-intensive, but can also result in inaccurate position labels. Semi-automatic capture methods have been devised where 3D joint locations are inferred from manually annotated 2D joint locations~\cite{oberweger2016efficiently}. Alternatives, which are still time-consuming, combine tracking a hand model and manual refinement, if necessary iterating these steps~\cite{sun2015cascade,tang2014latent,tompson2014real}. 
Additional sensors can aid automatic capture significantly, but care must be taken not to restrict the range of motion, and to minimise the depth appearance difference from bare hands, for example when using a data-glove~\cite{XuEtAl_IJCV15}.
More recently, less intrusive magnetic sensors have been employed for finger tip annotation in the {\it HandNet} dataset~\cite{wetzler2015rule} .

In this paper, we introduce our million-scale {\it BigHand2.2M} dataset that makes a significant advancement in terms of completeness of hand data variation and annotation quality, see Figure~\ref{pic:challengingposes} and Table~\ref{tab:ExistingBenchmarks}. We detail the capture set-up and methodology that enables efficient hand pose annotation with high accuracy. This enables us to capture the range of hand motions that can be adopted without external forces. Our dataset contains 2.2 million depth maps with accurately annotated joint locations.
The data is captured by attaching six magnetic sensors on the hand, five on each finger nail and one on the back of the palm, where each sensor provides accurate 6D measurements. Locations of all joints are obtained by applying inverse kinematics on a hand model with 31 degrees of freedom (dof) with  kinematic constraints.
The {\it BigHand2.2M} dataset also contains 290K frames of egocentric hand poses, which is 130 times more than previous egocentric hand pose datasets (Table~\ref{tab:egobenchmarkcomp}). Training a Convolutional Neural Network (CNN) on the data shows significantly improved results.
The recent study by Supancic \textit{et al.} \cite{supancic2015depth} on cross-benchmark testing showed that approximately 40\% of poses are estimated with an error larger than 50mm. This is due to a different capture set-up, hand shape variation, and annotation schemes.
Training a CNN using the {\it BigHand2.2M} dataset, we demonstrate state-of-the-art performance on existing benchmarks, 15-20mm average errors.

\vspace{-3mm}
\section{Existing benchmarks}
\label{para:relatedwork}
\vspace{-1mm}

\begin{table*}[t]
	\centering
	\resizebox{\textwidth}{!}{
		\begin{tabular}{llrrrrrr}
\toprule
    Dataset  & Annotation        &No. frames&  No. joints & No. subjects & View point   &Depth map resolution    \\
\midrule
    Dexter 1 \cite{handtracker_iccv2013}      & manual            & 2,137  &5  &1    & 3rd     & 320$\times$240                \\ 
    MSRA14 \cite{qian2014realtime}           & manual            & 2,400  &21 & 6   & 3rd     & 320$\times$240        \\
	  ICVL \cite{tang2014latent}               & track + refine    & 17,604 &16 & 10  & 3rd     & 320$\times$240      \\		
		NYU \cite{tompson2014real}               & track + refine    & 81,009 &36 & 2   & 3rd     & 640$\times$480 \\
		MSRA15 \cite{sun2015cascade}             & track + refine    & 76,375 &21 & 9   & 3rd     & 320$\times$240              \\
		UCI-EGO \cite{rogez2015first}            & semi-automatic         & 400    &26 & 2   & ego     & 320$\times$240               \\ 
		Graz16 \cite{oberweger2016efficiently}   & semi-automatic         & 2,166  &21 & 6   & ego     & 320$\times$240   \\
		ASTAR \cite{XuEtAl_IJCV15}               & automatic         & 870    &20 & 30  & 3rd     & 320$\times$240  \\	
		HandNet \cite{wetzler2015rule}           & automatic         & 212,928&6  & 10  & 3rd     & 320$\times$240              \\
		MSRC \cite{sharp2015accurate}            & synthetic         & 102,000&22 & 1   & 3rd     & 512$\times$424     \\
		{\bf BigHand2.2M}                            & automatic         & 2.2M   &21 & 10  & full    & 640$\times$480              \\
\bottomrule  
	  \end{tabular}}
	\caption{{\bf Benchmark comparison}. Existing datasets are limited in the number of frames, due to their annotation methods, either manual or semi-automatic. Our automatic annotation method allows collecting fully annotated depth images at frame rate. Our dataset is collected with the latest Intel RealSense SR300 camera \cite{intelSR300}, which captures depth images at $640 \times 480$-pixel resolution.}
	\label{tab:ExistingBenchmarks}
\end{table*}

Existing benchmarks for evaluation and comparison are significantly limited in scale (from a few hundred to tens of thousands), annotation accuracy, articulation, view point, and hand shape \cite{qian2014realtime,sharp2015accurate,sun2015cascade,tang2014latent,tompson2014real,wetzler2015rule,oberweger2016efficiently}. 
The bottleneck for building a large-scale benchmark using captured data is the lack of a rapid and accurate annotation method. Creating datasets by manual annotation \cite{qian2014realtime, handtracker_iccv2013} is labor-intensive and can result in inaccurate labels. These benchmarks are small in size, \eg MSRA14\cite{qian2014realtime} and Dexter 1 \cite{handtracker_iccv2013} have only 2,400 and 2,137 frames, respectively, making them unsuitable for large-scale training.
Alternative annotation methods, which are still labor-intensive and time-consuming, track a hand model and manually refine the results, if necessary they iterating these two steps~\cite{sun2015cascade,tang2014latent,tompson2014real}. The ICVL dataset \cite{tang2014latent} is one of the first benchmarks and it was annotated using 3D skeletal tracking \cite{melax2013dynamics} followed by manual refinement. 
However, its scale is small and limitations of the annotation accuracy have been noted in the literature \cite{sun2015cascade,oberweger2016efficiently}. The NYU dataset \cite{tompson2014real} is larger with a wider range of view points. Its annotations were obtained by model-based hand tracking on depth images from three cameras. Particle Swarm Optimization was used to obtain the final annotation. This method often drifts to incorrect poses, where manual correction is needed to re-initialize the tracking process. The MSRA15 dataset \cite{sun2015cascade} is currently the most complex in the area \cite{oberweger2016efficiently}. It is annotated in an iterative way, where an optimization method \cite{qian2014realtime} and manual re-adjustment alternate until convergence. The annotation also contains errors, such as occasionally missing finger and thumb annotations. This benchmark has a large view point coverage, but it has only small variation in articulation, capturing 17 base articulations and varying each of them within a 500-frame sequence.

Two small datasets were captured using semi-automatic annotation methods \cite{oberweger2016efficiently,rogez2015first}. The UCI-EGO dataset \cite{rogez2015first} was annotated by iteratively searching for the closest example in a synthetic set and subsequent manual refinement. The Graz16 dataset \cite{oberweger2016efficiently} was annotated by iteratively annotating visible joints in a number of key frames and automatically inferring the complete sequence using an optimization method, where the appearance as well as temporal, and distance constraints are exploited. However, it remains challenging to annotate rapidly moving hands. It also requires manual correction when optimization fails. This semi-automatic method resulted in a 2,166-frame annotated egocentric dataset, which is also insufficient for large-scale training.

Additional sensors can aid automatic capture significantly \cite{pons2011outdoor,von2016human,wetzler2015rule,XuEtAl_IJCV15}, but care must be taken not to restrict the range of motion. The ASTAR dataset \cite{XuEtAl_IJCV15} used a {\it ShapeHand} data-glove\cite{ShapeHand}, but wearing the glove influences the captured hand images, and to some extent hinders free hand articulation. In the works of \cite{pons2011outdoor,von2016human}, full human body pose esimtation was treated as a state estimation problem given magnetic sensor and depth data. More recently, less intrusive magnetic sensors have been used for finger tip annotation in the HandNet dataset \cite{wetzler2015rule}, which exploits a similar annotation setting as our benchmark with trakSTAR magnetic sensors \cite{trakSTAR}. However, this dataset only provides fingertip locations, not the full hand annotations. 

Synthetic data has been exploited for generating training data \cite{riegler2015framework,rogez2015understanding,xu2013efficient}, as well as evaluation \cite{sharp2015accurate}. Even though one can generate unlimited synthetic data, there currently remains a gap between synthetic and real data. Apart from differences in hand characteristics and the lack of sensor noise, synthetically generated images sometime produce kinematically implausible and unnatural hand poses, see Figure~\ref{pic:msrcgt}. The MSRC benchmark dataset \cite{sharp2015accurate} is a synthetic benchmark, where data is uniformly distributed in the 3D view point space. However, the data is limited in the articulation space, where poses are generated by random sampling from six articulations. 

\section{Full hand pose annotation}
\label{para:poseannotation}

\begin{figure}
\begin{center}
		\includegraphics[trim=7cm 0cm 9cm 1cm, clip=true, width=0.4\textwidth]{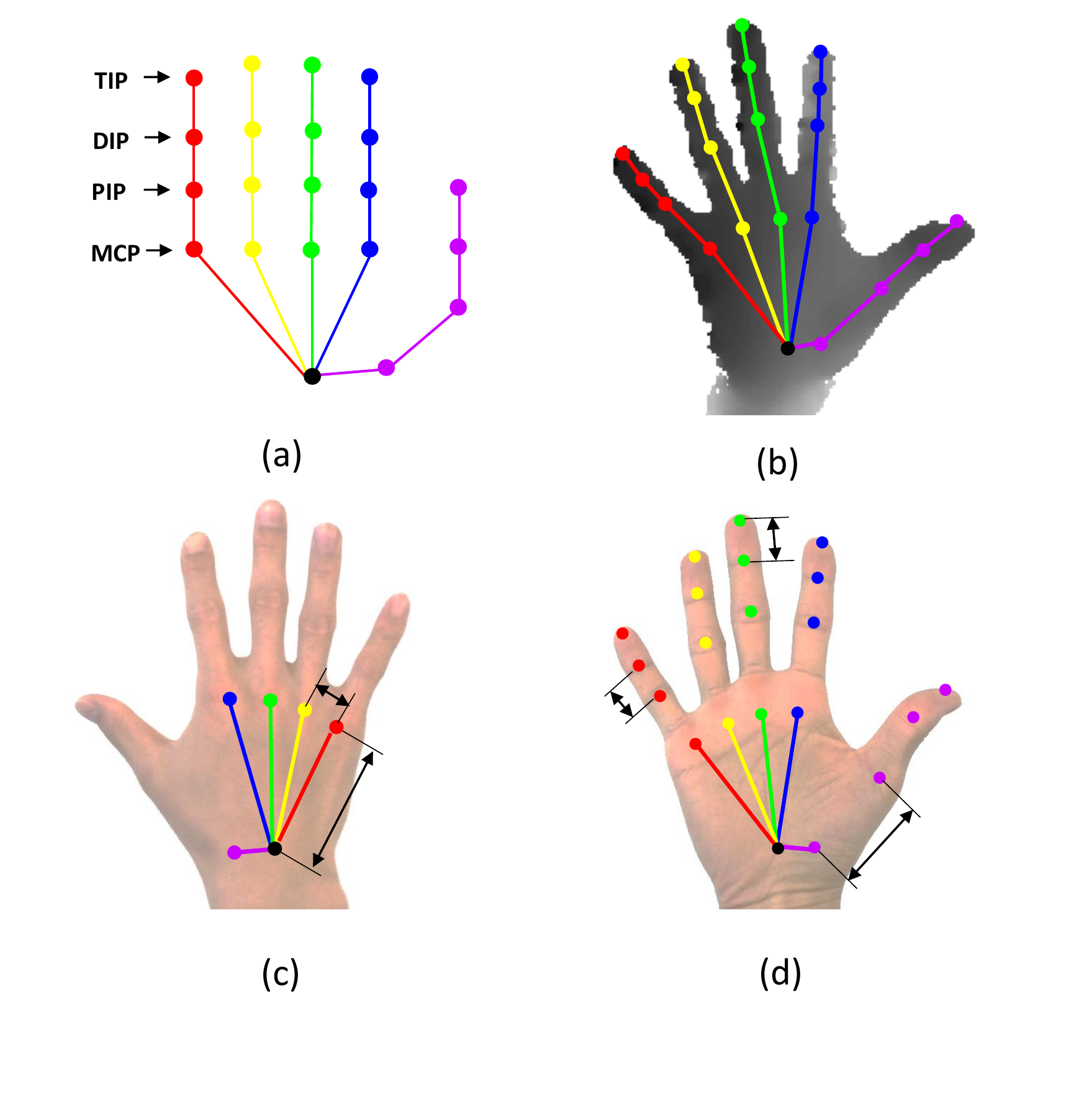}
		\vspace{-5mm}
    \caption{{\bf Hand model definition}. (a) Our hand model has 21 joints and moves with 31 degrees of freedom (dof). (b) Model fitted to hand shape. (c) and (d) show how hand shape is measured.}
    
	\label{pic:annotation_measurehand}
\end{center}
\end{figure}

\begin{figure}
\begin{center}
		\def\svgwidth{0.8\columnwidth}
		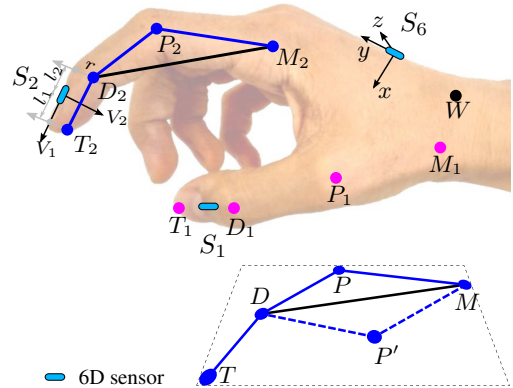
    \caption{{\bf Hand pose inference using six 6D magnetic sensors}. Global hand pose can be inferred from the location and orientation of sensor S6 on the back of the palm. Each sensor on the nail is used to infer the TIP and DIP joints of the corresponding finger. Each PIP joint can be calculated using bone lengths and physical constraints.}

	\label{pic:annotation_IK}
\end{center}
\end{figure}

\begin{figure}
\begin{center}
				\def\svgwidth{\columnwidth}
		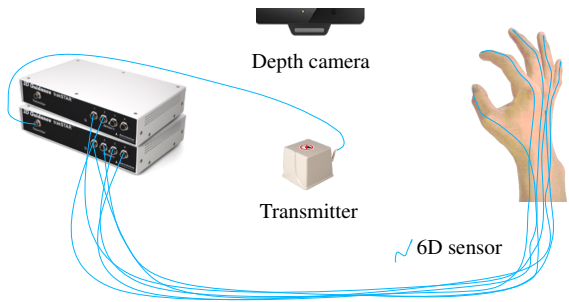
    \caption{{\bf Annotation settings}. The equipment used in our annotation system are: Two hardware synchronized electromagnetic tracking units, six 6D magnetic sensors, one mid-range transmitter, and an Intel RealSense SR300 camera.}

	\label{pic:annotation_equipments}
\end{center}
\end{figure}

In this section we present our method to do accurate full hand pose annotations using the trakSTAR tracking system with 6D magnetic sensors.

\subsection{Annotation by inverse kinematics}
\label{para:handposannot}

Our hand model has 21 joints and can move with 31 degrees of freedom (dof), as shown in Figure \ref{pic:annotation_measurehand}. We capture 31 dimensions, including 6 dimensions for global pose and 25 joint angles. Each finger's pose is represented by five angles, including the twist angle, flexion angle, abduction angle for the MCP joint and flexion angles for the DIP and PIP joints. For each subject, we manually measure the bone lengths, see Figure \ref{pic:annotation_measurehand} (c) and (d).
 
Given the six magnetic sensors, each with 6D dof (location and orientation), along with a hand model, we use inverse kinematics to infer the full hand pose defined by the locations of 21 joints, as shown in Figure \ref{pic:annotation_measurehand}. The physical constraints per subject are (1) the wrist and 5 MCP joints are fixed relative to each other, (2) bone lengths are constant, and (3) MCP, PIP, DIP, and TIP joints for each finger lie on the same plane.

Similar to \cite{schaffelhofer2012new} five magnetic sensors (from thumb to little finger, the sensors are S1, S2, S3, S4, S5) are attached on the five fingers' tips. The sixth sensor (S6) is attached to the back of the palm, see Figure~\ref{pic:annotation_IK}. Given the location and orientation of S6, as well as the hand model shape, the wrist (W) and five MCP joints (M1, M2, M3, M4, M5) are inferred. For each finger, given the sensor's location and orientation, the TIP and DIP are calculated in the following way (as shown in Figure~\ref{pic:annotation_IK}, take the index finger as an example): the sensor's orientation is used to find the three orthogonal axes, $V_{1}$ is along the finger, $V_{2}$ is pointing forward from the finger tip. The TIP (T) and DIP (D) joint locations are calculated as: 
\begin{equation}
T = L(S) + l_{1} V_{1} + r V_{2} ,
\end{equation}
\begin{equation}
D = L(S) - l_{2} V_{1} + r V_{2} ,
\end{equation}
where $L(S)$ denotes the sensor location, $r$ is half the finger thickness, and $l_{1}$+$l_{2}$ = $b$, where $b$ is the bone length connecting the DIP and TIP joints.
The final joint to infer is the PIP, shown at location P in Figure~\ref{pic:annotation_IK}, is calculated using the following conditions: (1) T, M, D are given, (2) $\Vert{P-D}\Vert$ and $\Vert{P-M}\Vert$ are fixed, (3) T, D, P, M are on the same plane, and (4) T and P should be on different sides of the line connecting M and D. These constraints are sufficient to uniquely determine P.

\subsection{Synchronization and calibration}
\label{para:equipments}

To build and annotate our dataset, we use a trakSTAR tracking system \cite{trakSTAR} combined with the latest generation Intel RealSense SR300 depth sensor \cite{intelSR300}, see Figure~\ref{pic:annotation_equipments}. The trakSTAR system consists of two hardware synchronized electromagnetic tracking units, each of which can track up to four 6D magnetic sensors. The 6D sensor (``Model 180") is 2mm wide and is attached to a flexible 1.2mm wide and 3.3m long cable. When the cable is attached to the hand using tight elastic loops the depth images and hand movements are minimally affected. We use the mid-range transmitter with a maximum tracking distance of 660mm, which is suitable for hand tracking. The tracking system captures the locations and orientations of the six sensors at 720fps and is stable and without drift in continuous operation.
The depth camera captures images with a resolution of $640 \times 480$  and runs at a maximum speed of 60fps. The measurements are synchronized by finding the nearest neighboring time stamps. The time gap between the depth image and the magnetic sensors in this way is 0.7 millisecond at most. 

The trakStar system and the depth sensor have their own coordinate systems, and we use a solution to the perspective-N-point problem to calibrate the coordinates as in \cite{wetzler2015rule}. Given a set of 3D magnetic sensor locations and the corresponding 2D locations in the depth map as well as intrinsic camera parameters, the ASPnP algorithm \cite{zheng2013aspnp}  estimates the transformation between these two coordinate systems.

\section{BigHand2.2M benchmark}
\label{para:postproc_neat}

\begin{figure*}[t]
\begin{center}
\includegraphics[trim=2.8cm 1.5cm 1cm 1cm, clip=true, width=0.3\textwidth, height=0.29\textwidth]{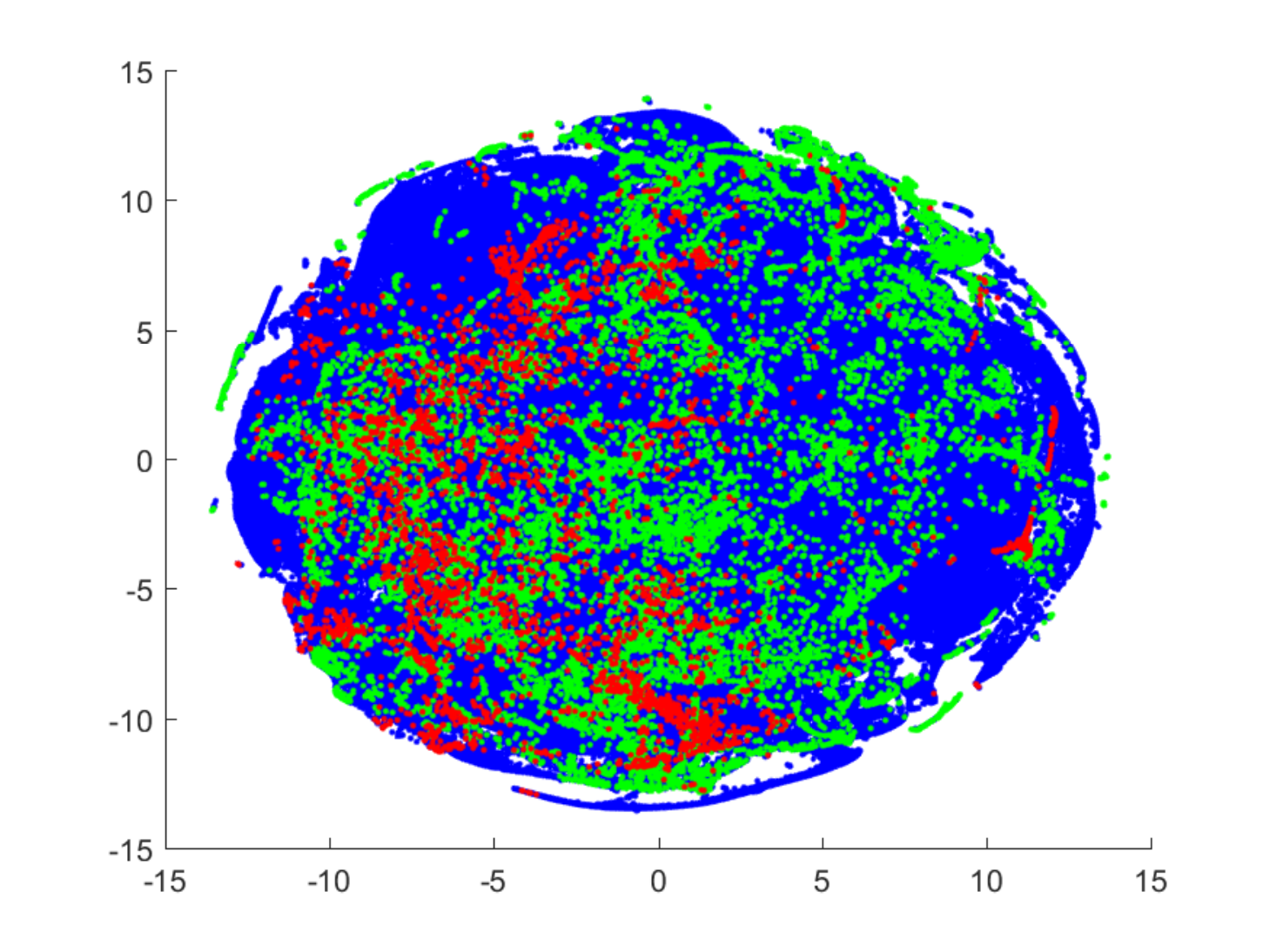}
    \includegraphics[trim=7cm 3cm 3cm 3cm, clip=true, width=0.3\textwidth, height=0.28\textwidth]{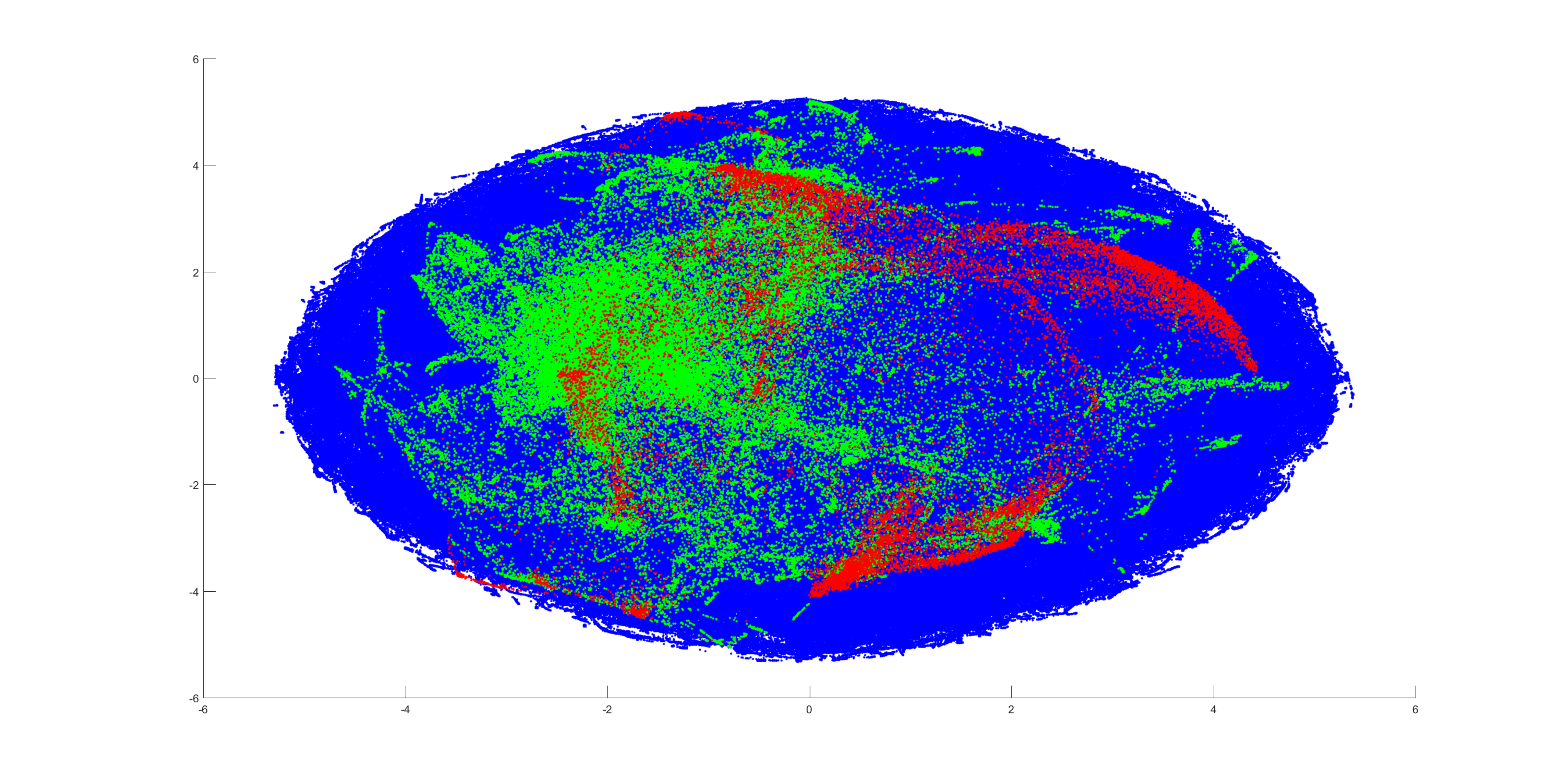}
		\includegraphics[trim=7cm 3cm 3cm 3cm, clip=true, width=0.3\textwidth, height=0.28\textwidth]{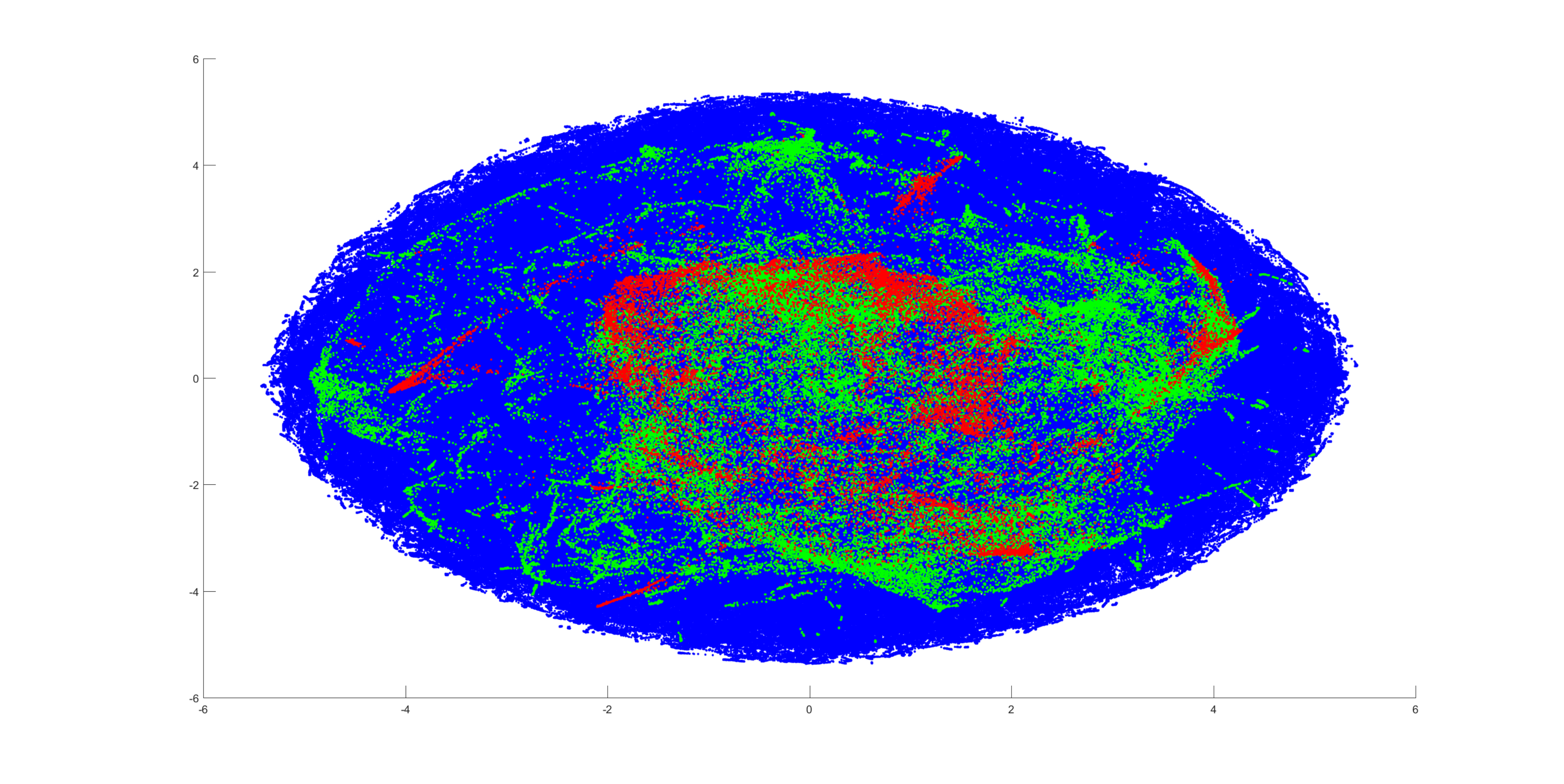} \hfill \\	
                
		\caption{{\bf 2D t-SNE embedding of the hand pose space}. \textcolor[rgb]{0,0,1}{{\it BigHand2.2M}} is represented by blue, \textcolor[rgb]{1,0,0}{ICVL} by red, and \textcolor[rgb]{0,1,0}{NYU} by green dots. The figures show (left) global view point space coverage, (middle) articulation space (25D), and (right) combined of global orientation and articulation coverage. Compared with existing datasets, the {\it BigHand2.2M} contains a more complete range of variation.}
	\label{pic:dataset}
\end{center}
\end{figure*}

We collected the {\it BigHand2.2M} dataset containing 2.2 million depth images of single hands with annotated joints (see Section~\ref{para:poseannotation}). Ten subjects (7 male, 3 female) were captured for two hours each.

\begin{table}
	\centering
	\resizebox{\columnwidth}{!}{
\begin{tabular}{lccc} 
\toprule
    Benchmarks   &  Rogez~\cite{rogez2015first}& Oberweger~\cite{oberweger2016efficiently}   &  BigHand2.2M Egocentric  \\	
\midrule
	  No. Frames       &  400                     & 2166         & 290K            \\
	\bottomrule			
\end{tabular}}
	\caption{Egocentric Benchmark size comparison. The egocentric subset of {\it BigHand2.2M} dataset is 130 time larger than the next largest available dataset.}
	\label{tab:egobenchmarkcomp}
\end{table}

\subsection{Hand view-point space}
\label{para:viewpointspace}

In order to cover diverse view points, we vary the sensor height, the subject's position and arm orientation.
The view point space (a hemisphere for the 3rd person view point) is divided into 16 regions (4 regions uniformly along each of two 3D rotation axes), and subjects are instructed to carry out random view point changes within each region. In addition, our dataset collects random changes in the egocentric view point.
As the t-SNE visualization in Figure~\ref{pic:dataset}(left) shows, our benchmark data covers a significantly larger region of the global view-point space than the ICVL and NYU dataset.

\subsection{Hand articulation space}
\label{para:ariculationspace}

Similar to \cite{wu2001capturing}, we define 32 {\it extremal poses} as hand poses where each finger assumes a maximally bent or extended position. For maximum coverage of the articulation space, we enumerate all ${32 \choose 2}$ = 496 possible pairs of these {\it extremal poses}, and capture the natural motion when transitioning between the two poses of each pair.

In total the {\it BigHand2.2M} dataset consists of three parts:
(1) Schemed poses:  to cover all the articulations that a human hand can freely adopt, this contains has 1.534 million frames, captured as described above.
(2) Random poses: 375K frames are captured with participants being encouraged to fully explore the pose space.
(3) Egocentric poses: 290K frames of egocentric poses are captured with subjects carrying out the 32 {\it extremal poses} combined with random movements.

As Figure~\ref{pic:dataset} (middle, right) shows, our benchmark spans a wider and denser area in the articulation and the combined of articulation and view-point space, compared to the ICVL and NYU.

\subsection{Hand shape space} 
\label{para:handshapespace}

\begin{figure}[ht]
\begin{center}
    \includegraphics[width=0.48\textwidth, height=0.25\textwidth]{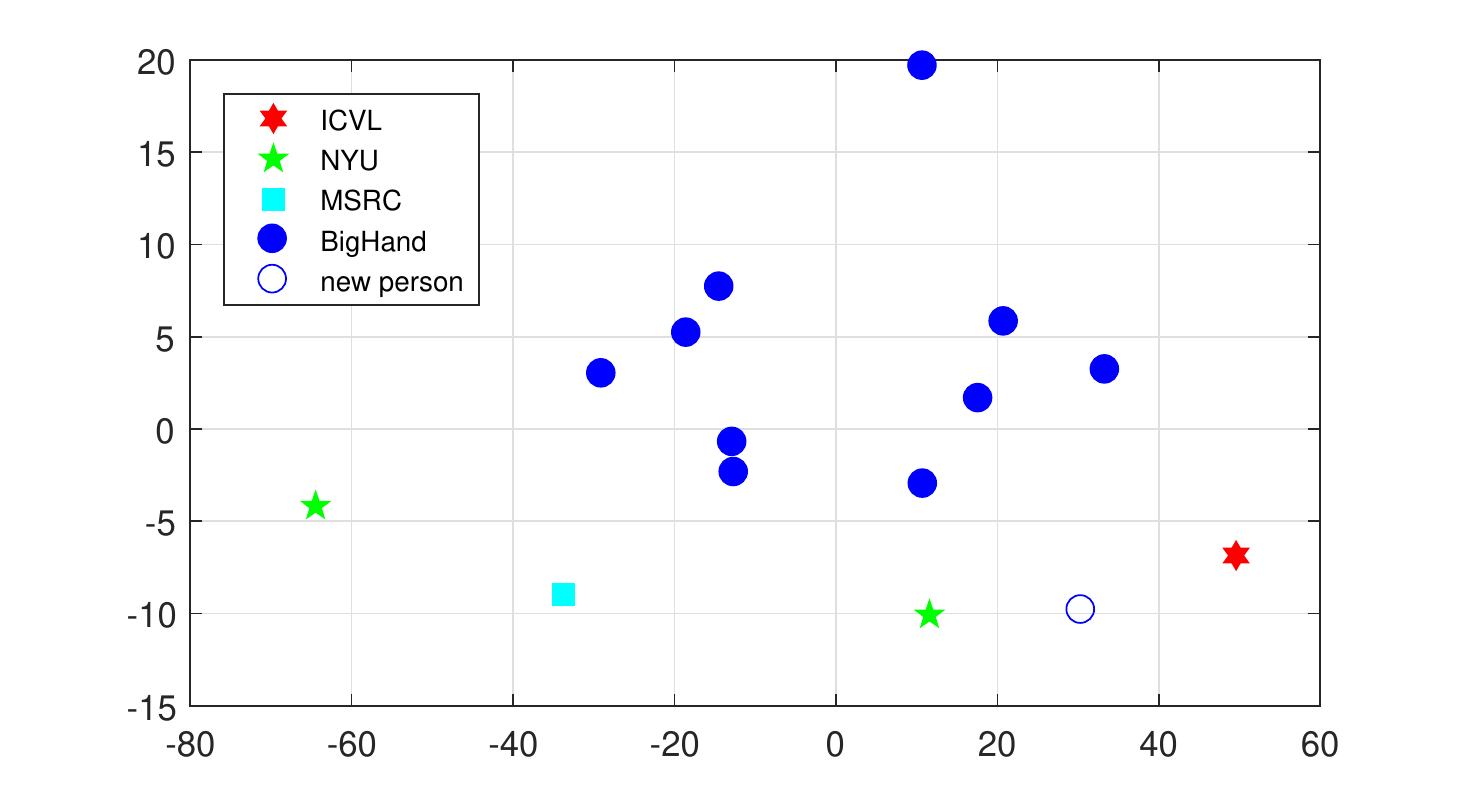}
    \caption{{\bf Hand shape variation}. Hand variation is visualized by applying PCA to the shape parameters. The {\it BigHand2.2M}  dataset contains 10 hand shapes and an additional shape for testing. The ICVL dataset contains one hand shape due to its annotation method. The NYU dataset includes two hand shapes, the MSRC dataset includes one synthetic hand shape.} 

	\label{pic:handshape_comparison}
\end{center}
\end{figure}

We select ten participants with different hand shapes (7 male, 3 female, age range: 25-35 years).
Existing benchmarks also use different participants, but are limited in annotated hand shapes due to annotation methods. Figure~\ref{pic:handshape_comparison} visualizes shapes in different datasets using the first two principal components of the hand shape parameters. The ICVL dataset \cite{tang2014latent} includes ten participants with similar hand size, and all are annotated with a single hand shape model. The NYU \cite{tompson2014real} training data uses one hand shape, while its test data uses two hand shapes,  one of which is from the training set. The MSRA15 dataset includes nine participants, but in the annotated ground truth data, only three hand shapes are used. The MSRC \cite{sharp2015accurate} synthetic benchmark includes a single shape. 

In the experiments, we use the dataset of 10 subjects for training, and testify how well the learnt model generalises to different shapes in existing benchmarks (cross-benchmark) and an unseen new shape (``new person" in Figure~\ref{pic:handshape_comparison}). See section~\ref{para:stateoftheartanalysis} for more explanations. 

\begin{figure*}[h]
\begin{center}
    \includegraphics[width=0.48\textwidth]{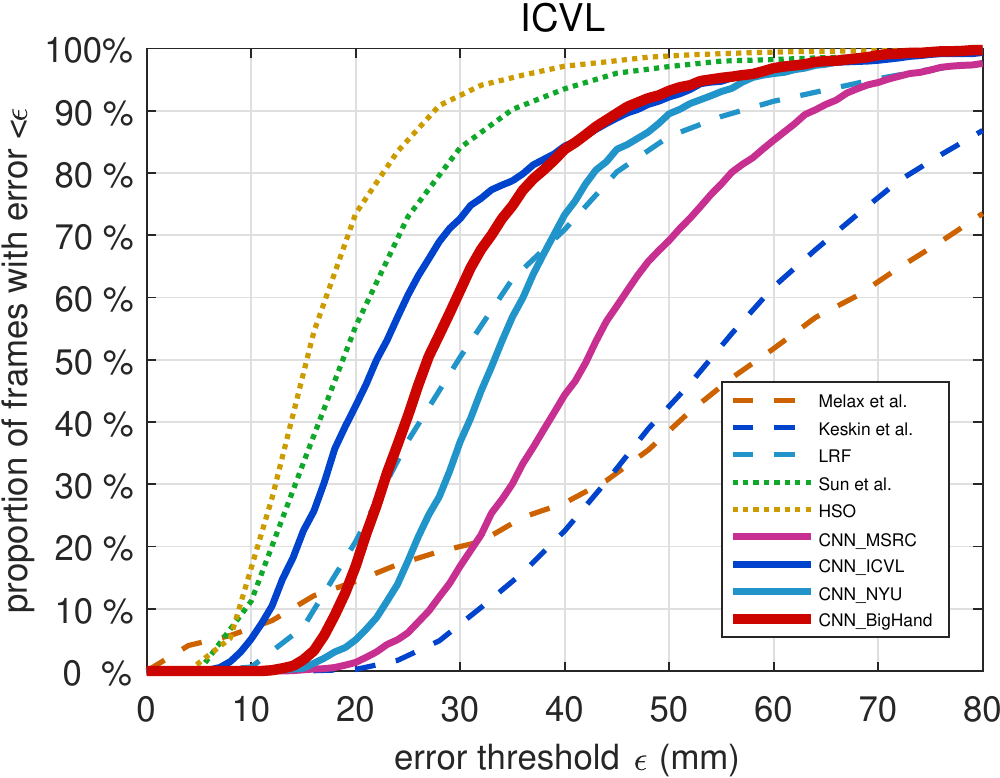}\hfill
		\includegraphics[width=0.48\textwidth]{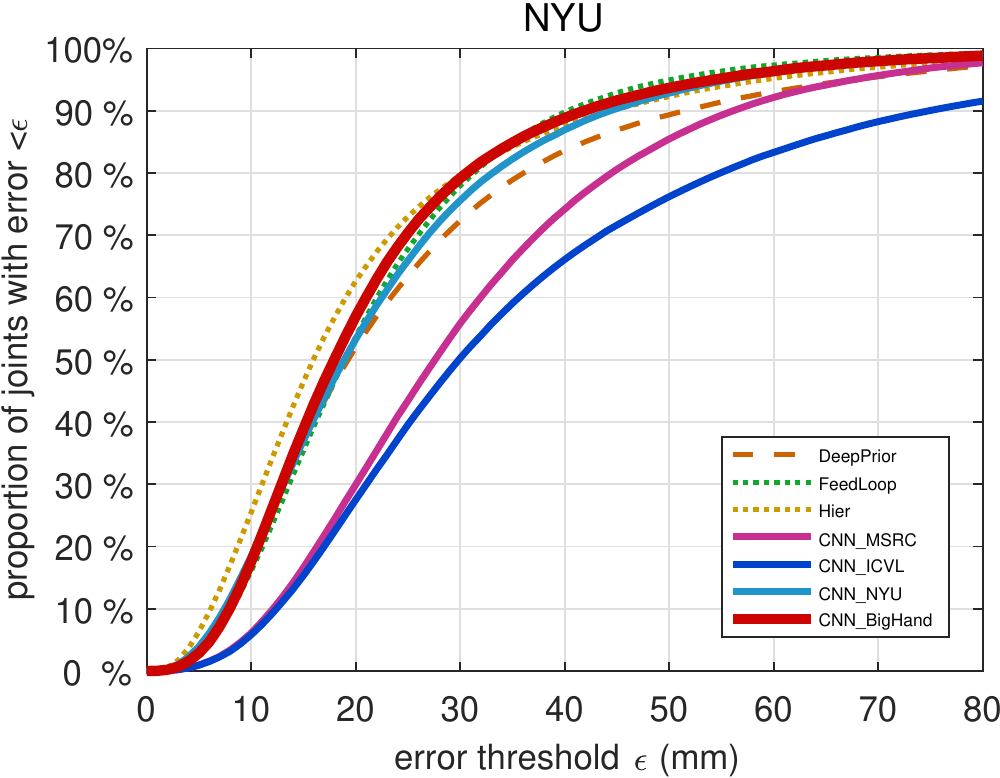}\\
    \caption{{\bf Cross-benchmark performance}. CNN models are trained on the ICVL, NYU, MSRC, and the new {\it BigHand2.2M} dataset, respectively, and evaluated on (left)  ICVL and (right) NYU test data. A CNN trained on {\it BigHand2.2M} achieves state-of-the-art performance on ICVL and NYU, while the CNNs trained on ICVL, NYU, and MSRC do not generalize well to other benchmarks.  The networks CNN\_MSRC, CNN\_ICVL, CNN\_NYU, and CNN\_BigHand are trained on the training set of MSRC, ICVL, NYU, and {\it BigHand2.2M}, respectively.}
	\label{pic:crossbenchmark}
\end{center}
\end{figure*}

\section{Analysis of the state of the art}
\label{para:stateoftheartanalysis}

In this section we use the {\it Holi} CNN architecture~\cite{ye2016spatial} as the current state of the art. The detailed structure is shown in the supplementary material. The input for the CNN model is the cropped hand area using the ground truth joint locations. This region is normalized to 96 $\times$ 96 pixels and is fed into the CNN, together with two copies downsampled to 48 $\times$ 48 and 24 $\times$ 24. The cost function is the mean squared distance between the location estimates and the ground truth locations.
The CNN is implemented using Theano and is trained on a desktop with an Nvidia GeForce GTX TITAN Black and a 32-core Intel processor. The model is trained using Adam, with $\beta_1=0.9$, $\beta_2=0.999$ and $\alpha=0.0003$. We stop the training process when the cost of the validation set reaches the minimum,  where each training epoch takes approximately 40 minutes. When training the CNN model on {\it BigHand2.2M}, ICVL, NYU and MSRC, we keep the CNN structure and $\beta_1$, $\beta_2$, $\alpha$ of Adam unchanged.

All frames of 10 subjects are uniformly split into a training set and a validation set with a 9-to-1 ratio, which is similar to ICVL, NYU, and HandNet \cite{wetzler2015rule}. In addition to the 10 subjects, a challenging test sequence of 37K frames of a new subject is recorded and automatically annotated, as shown ``new person" in Figure~\ref{pic:handshape_comparison}. For a quantitative comparison, we measure the ratio of joints within a certain error bound $\epsilon$ \cite{ye2016spatial,tang2015opening,sharp2015accurate}.

\begin{figure}[t]
\begin{center}
\includegraphics[width=0.45\textwidth]{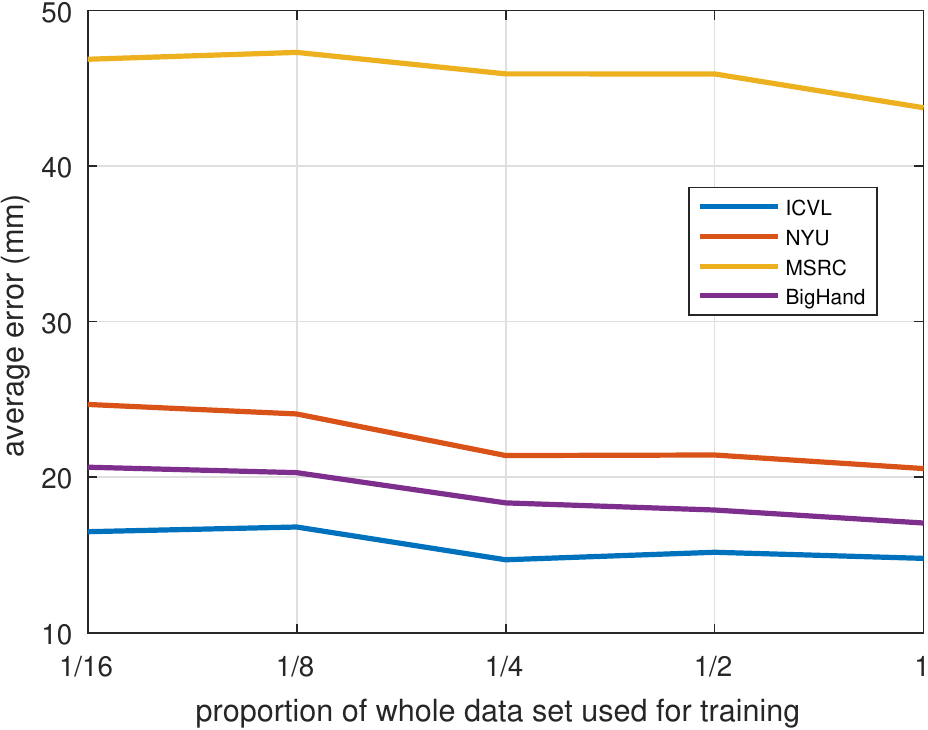}\hfill \\
\caption{{\bf Data size effect on cross benchmark evaluation.} When the CNN model is trained on $\frac{1}{16}$, $\frac{1}{8}$, $\frac{1}{4}$, $\frac{1}{2}$, and all of the benchmark data, the test results on ICVL, NYU, MSRC, and {\it BigHand2.2M} keep improving.}

	\label{pic:datasizeeffect}
\end{center}
\end{figure}

\begin{table}
                \centering
                \resizebox{\columnwidth}{!}{
\begin{tabular}{lrrrr}

\toprule
       \backslashbox{train}{test}&  ICVL & NYU   & MSRC & {\it BigHand2.2M} \\
\midrule
                  ICVL       &  \textbf{12.3}                     & 35.1         & 65.8 &   46.3        \\
                  NYU       &  20.1                      & \textbf{21.4}        & 64.1 &  49.6        \\
                  MSRC       &  25.3                    & 30.8        & \textbf{21.3} &  49.7       \\
                  {\it BigHand2.2M}       &  \textbf{14.9}                     & \textbf{20.6}      &       \textbf{43.7} &  \textbf{17.1}        \\
                \bottomrule                                      
\end{tabular}}
                \caption{{\bf Cross-benchmark comparison}. Mean errors of CNNs trained on ICVL, NYU, MSRC and {\it BigHand2.2M} when cross-tested. The model trained on {\it BigHand2.2M} performs well on ICVL and NYU, less so on the synthetic MSRC data. Training on ICVL, NYU, or MSRC does not generalize well to other datasets.}
                \label{tab:crossbenchmark}
\end{table}

\begin{figure}[t]
	\centering
    \includegraphics[trim=9.5cm 4cm 10.5cm 4cm, clip=true,width=0.09\textwidth]{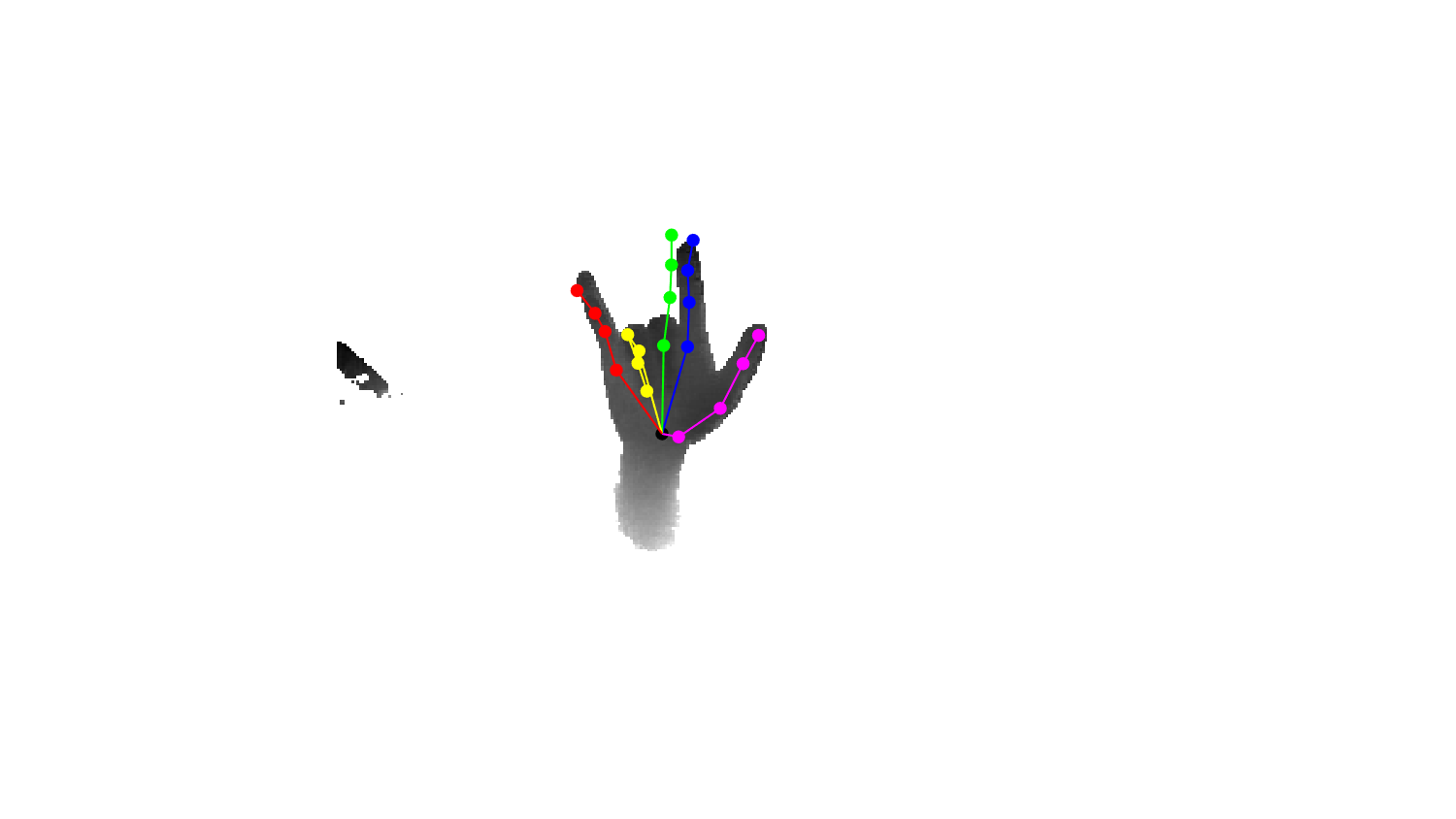}
	\includegraphics[trim=9.5cm 4cm 10.5cm 4cm, clip=true,width=0.09\textwidth]{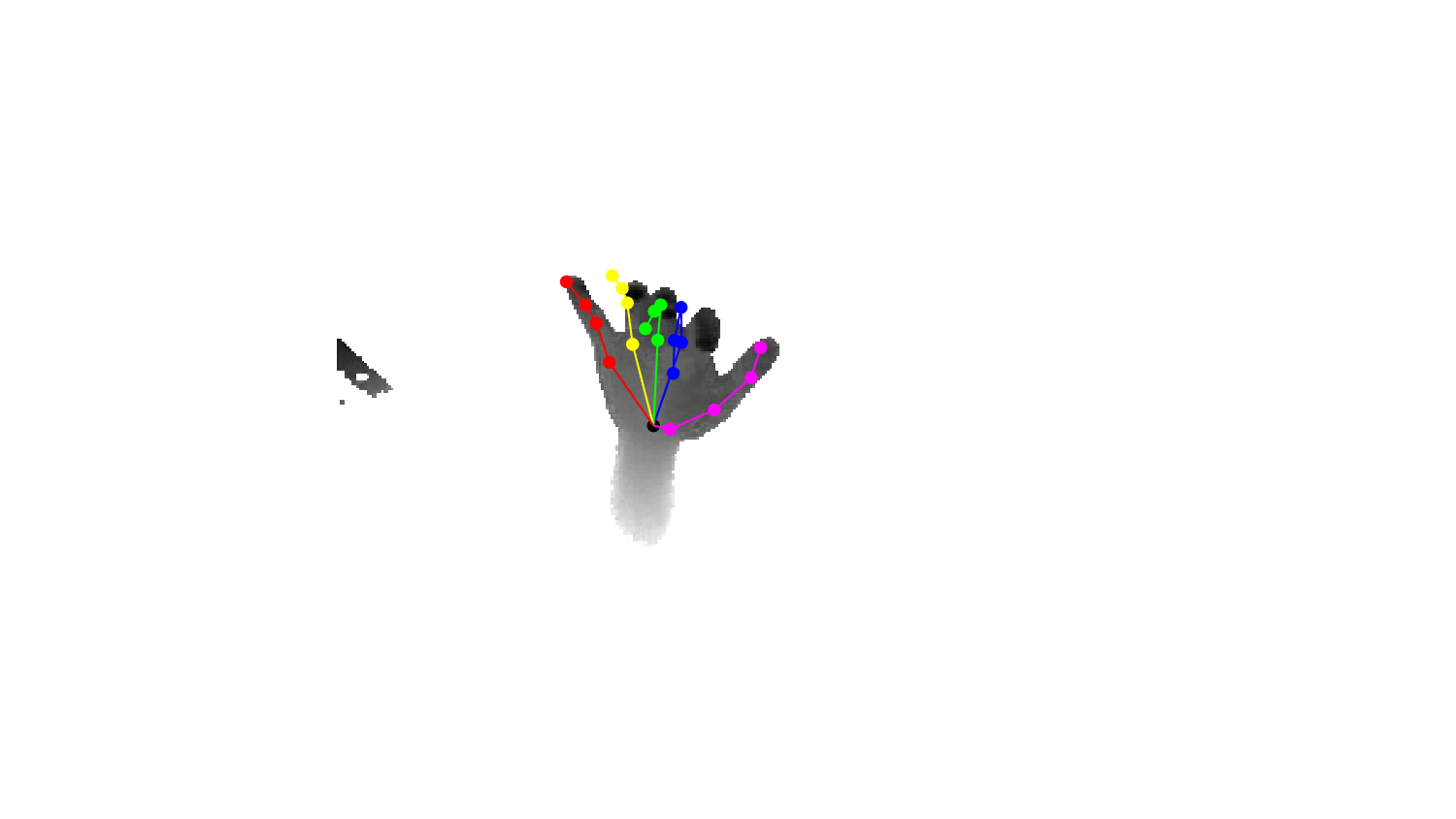}	
	\includegraphics[trim=10cm 4cm 10cm 4cm, clip=true,width=0.09\textwidth]{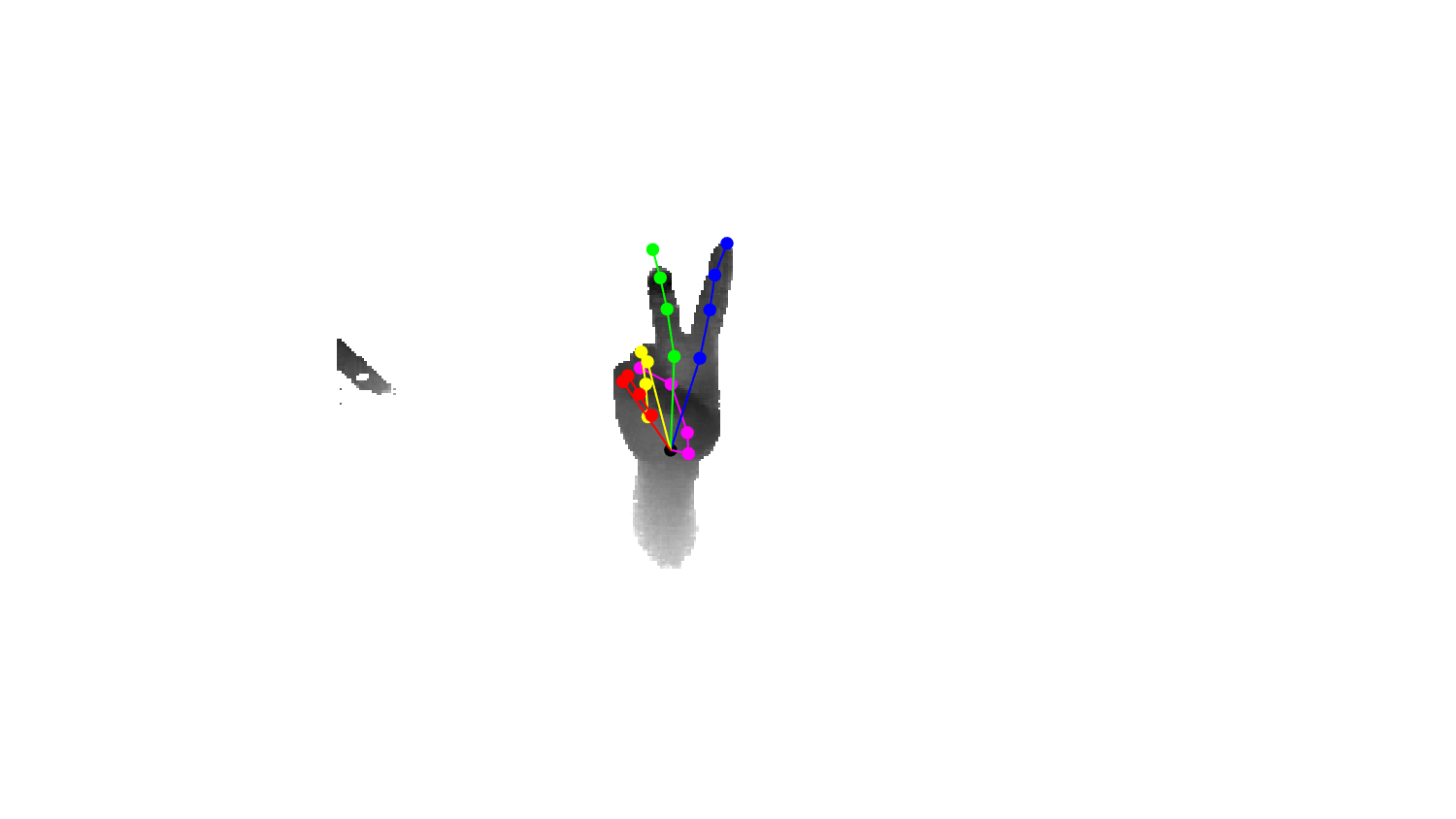}
	\includegraphics[trim=10.5cm 4.7cm 9.5cm 3.3cm, clip=true,width=0.09\textwidth]{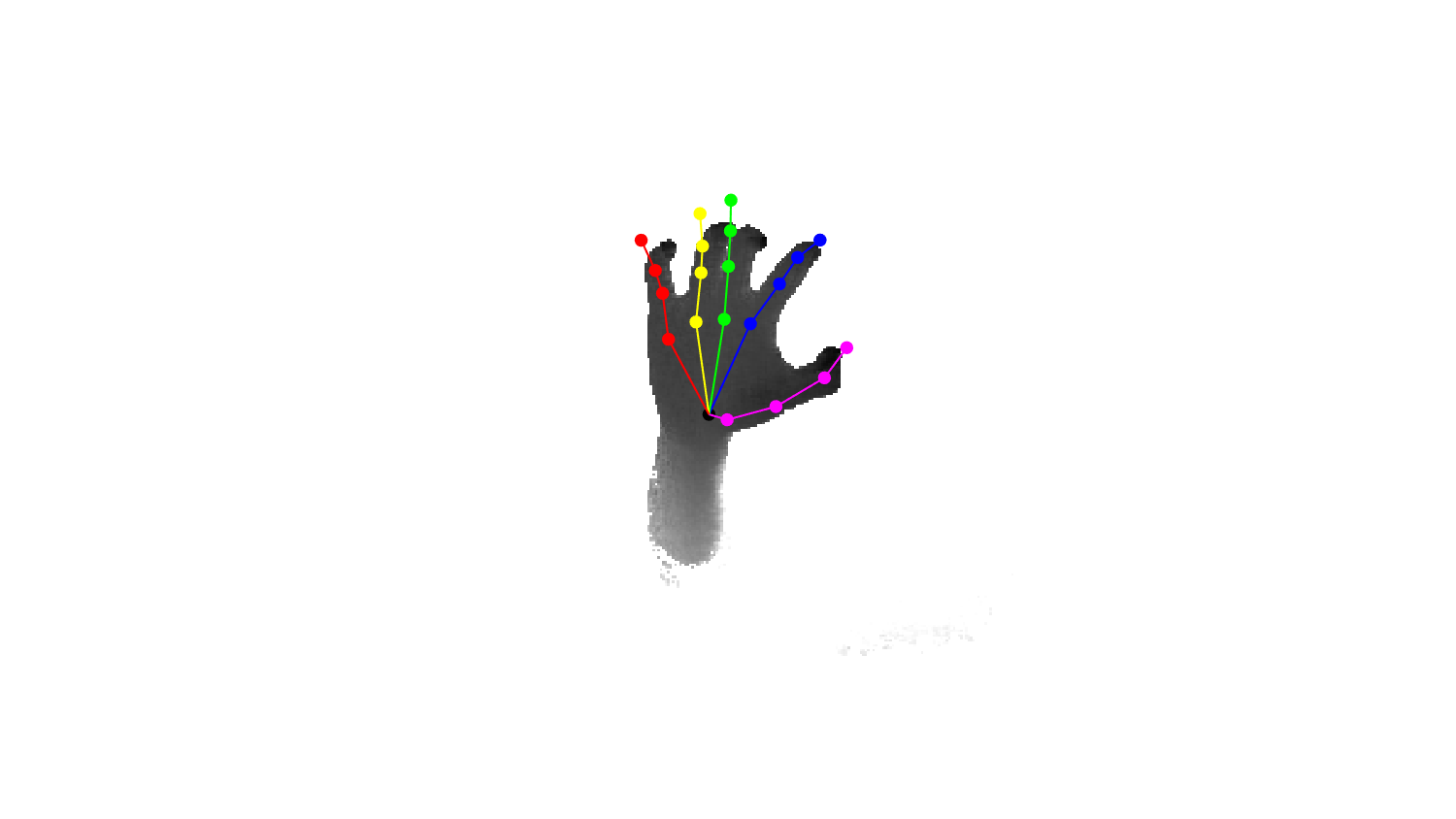}
	\includegraphics[trim=10cm 4.5cm 10cm 3.5cm, clip=true,width=0.09\textwidth]{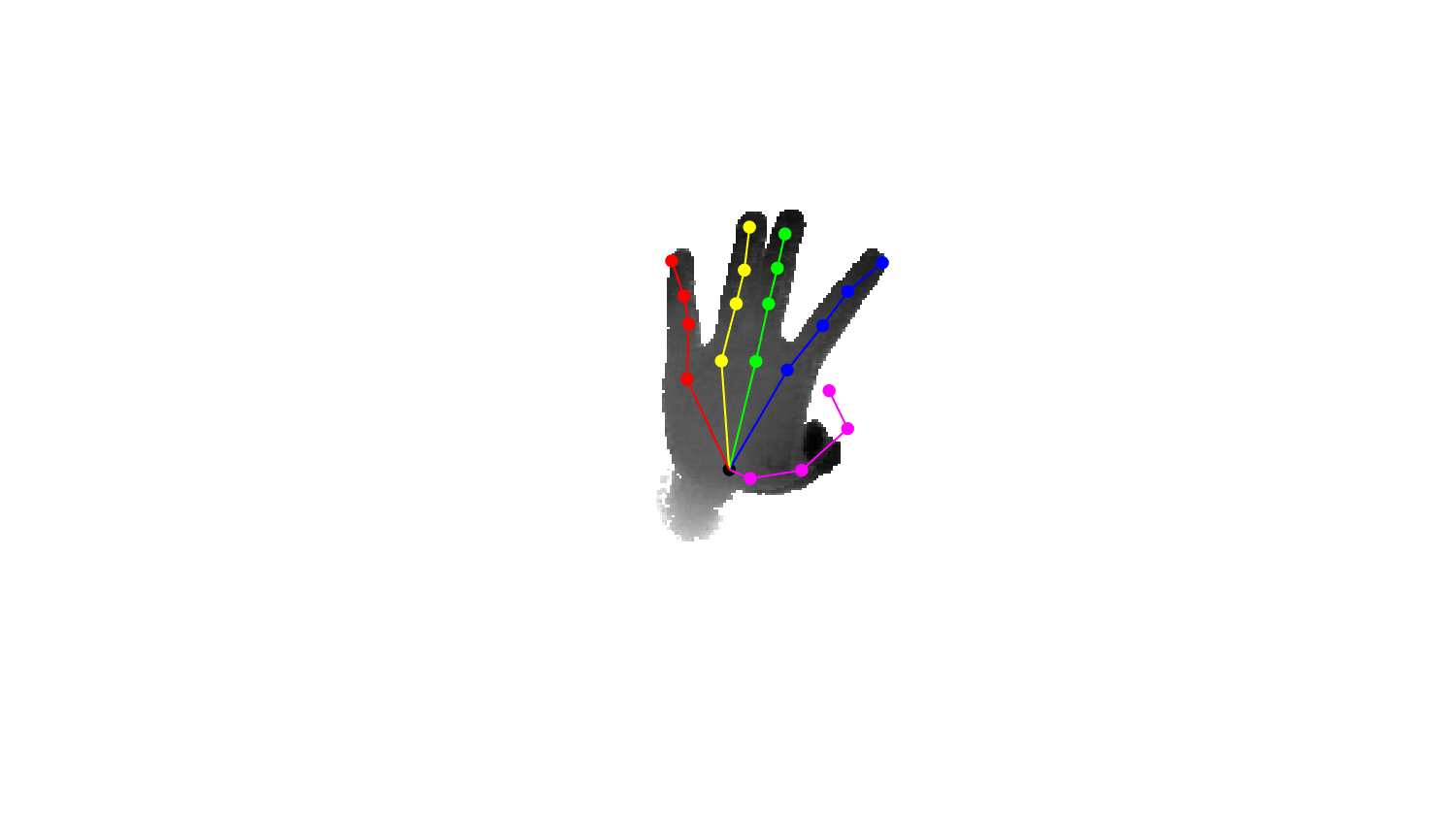}\hfill \\	

	\includegraphics[trim=9.5cm 4cm 10.5cm 4cm, clip=true,width=0.09\textwidth]{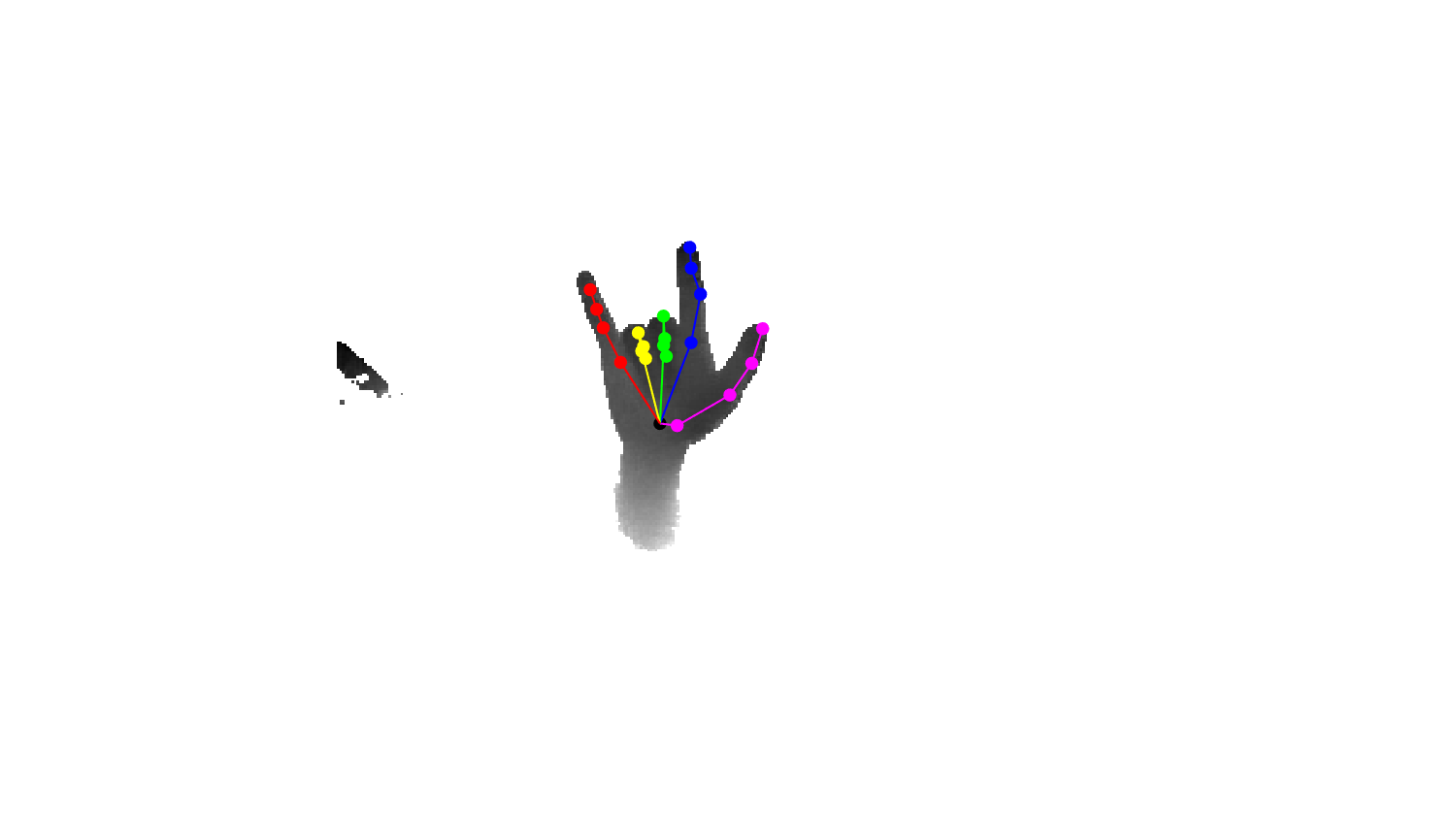}
	\includegraphics[trim=9.5cm 4cm 10.5cm 4cm, clip=true,width=0.09\textwidth]{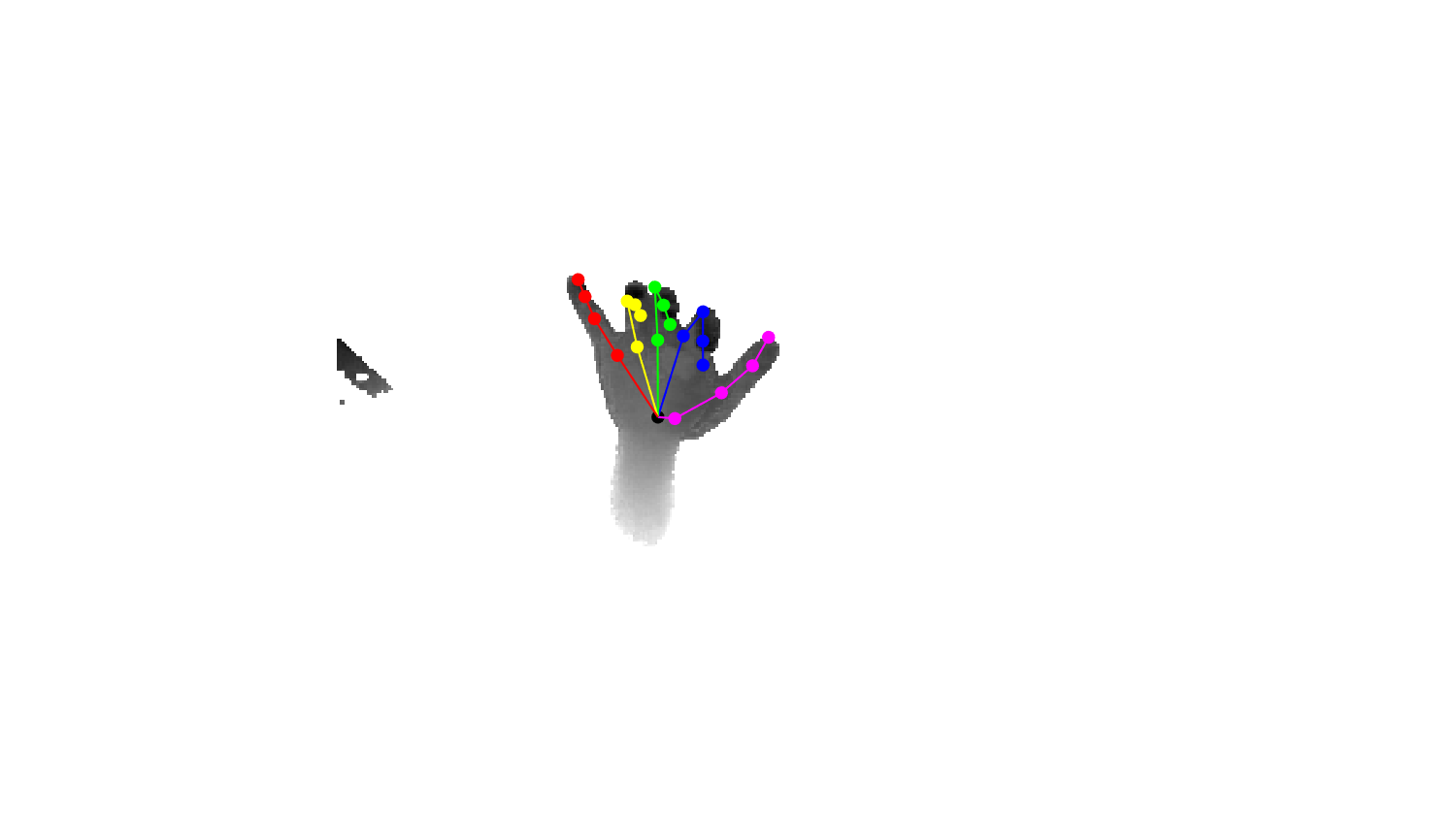}	
	\includegraphics[trim=10cm 4cm 10cm 4cm, clip=true,width=0.09\textwidth]{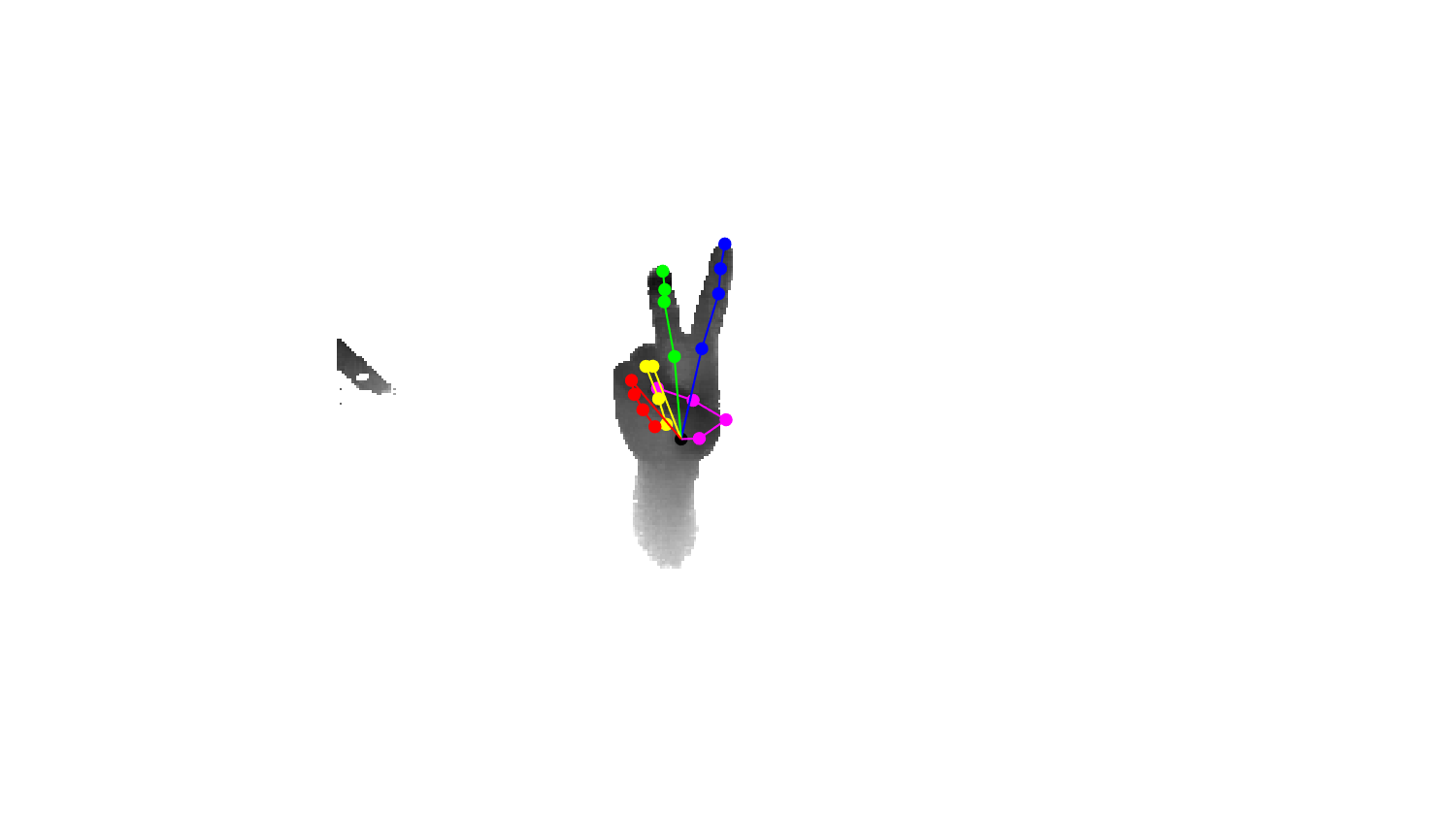}
	\includegraphics[trim=10.5cm 4.7cm 9.5cm 3.3cm, clip=true,width=0.09\textwidth]{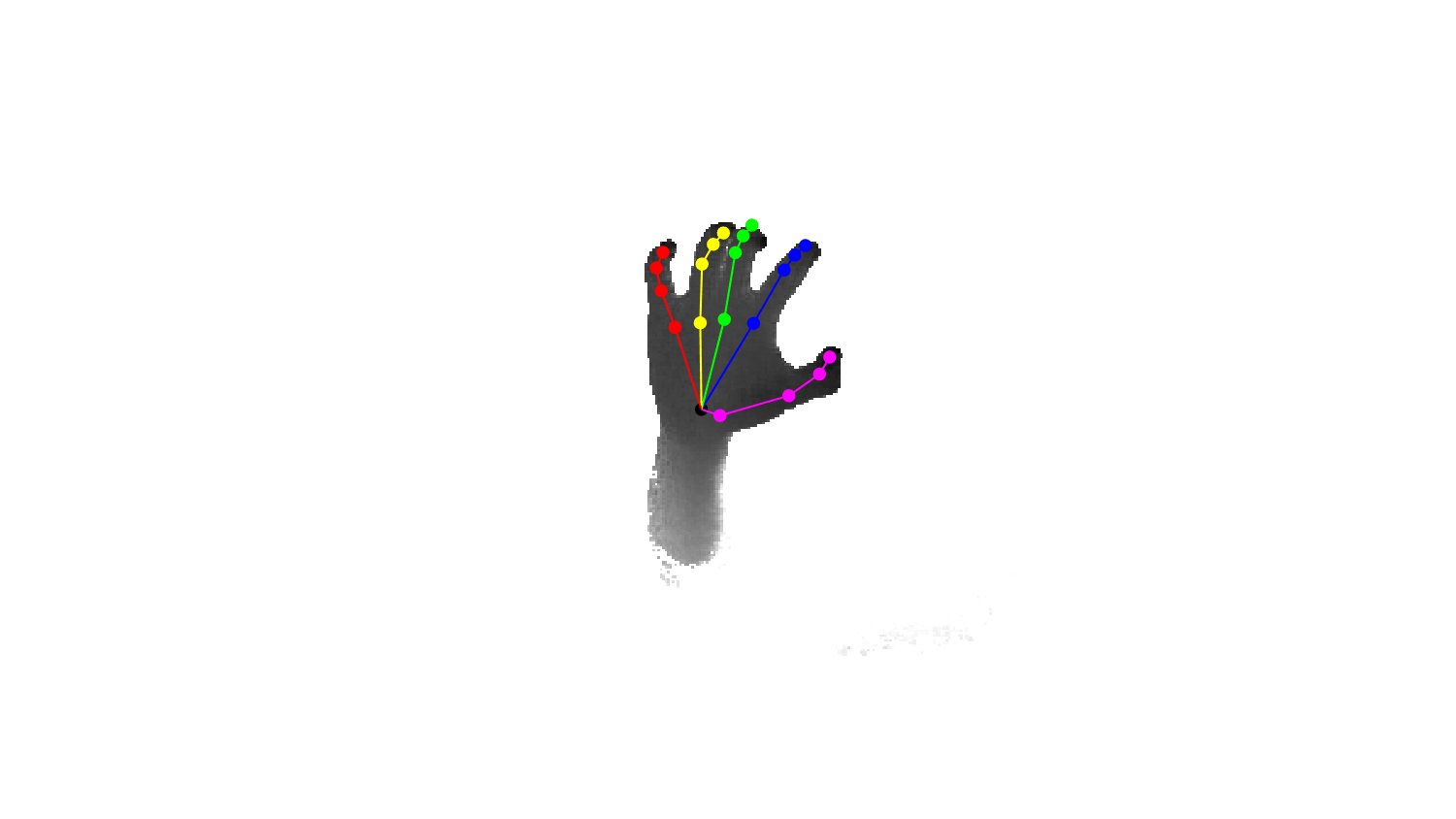}
	\includegraphics[trim=10cm 4.5cm 10cm 3.5cm, clip=true,width=0.09\textwidth]{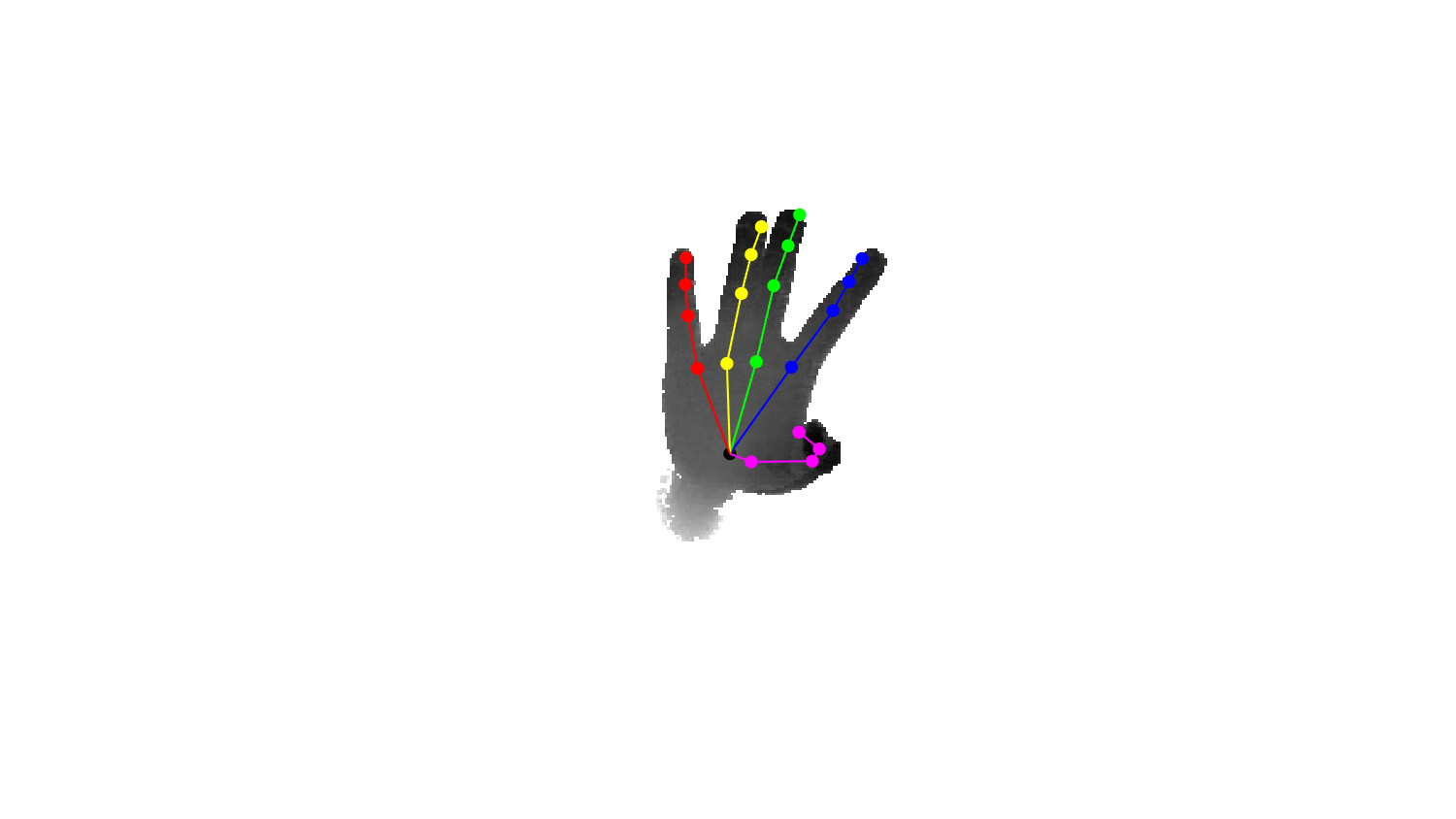}\hfill \\	
	
	 \caption{{\bf Generalization of the CNN trained on {\it BigHand2.2M}}. The CNN generalizes to the  ICVL dataset with a lower error than the original annotated ground truth. (top) ICVL ground truth annotations, (bottom) our estimation results.}

     \label{pic:icvlgt}
\end{figure}

\begin{figure}[t]
	\centering

	 \includegraphics[trim=3cm 1cm 2cm 1cm, clip=true,width=0.09\textwidth]{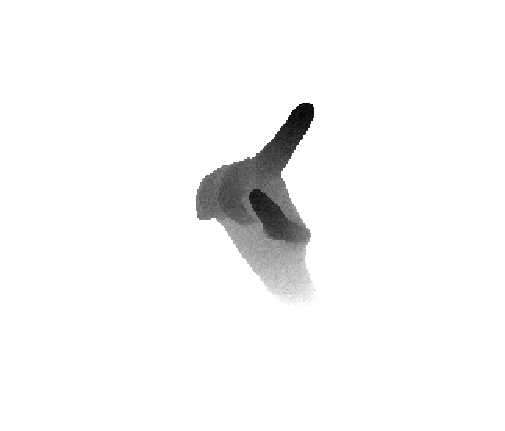}
   \includegraphics[trim=2.5cm 0.3cm 2.5cm 1.7cm, clip=true,width=0.09\textwidth]{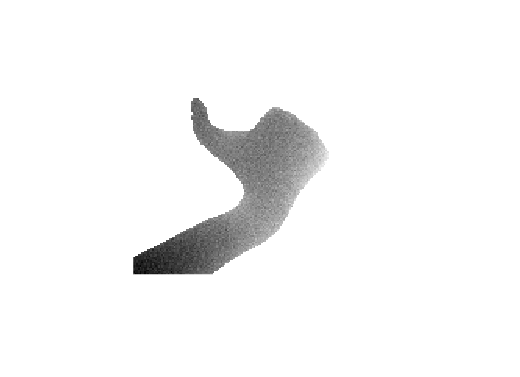}
   \includegraphics[trim=2.3cm 1.5cm 2.7cm 0.5cm, clip=true,width=0.09\textwidth]{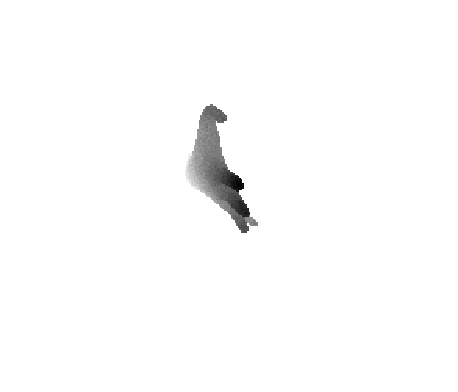}
	 \includegraphics[trim=3cm 1.5cm 3cm 0.5cm, clip=true,width=0.09\textwidth]{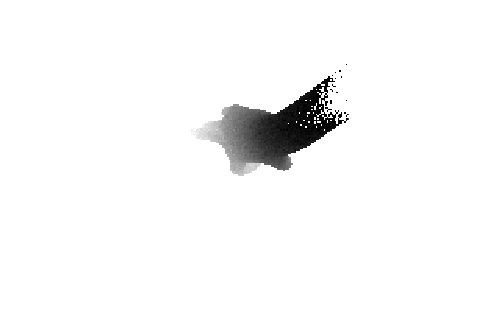}
	 \includegraphics[trim=2.5cm 0.5cm 2.5cm 1.5cm, clip=true,width=0.09\textwidth]{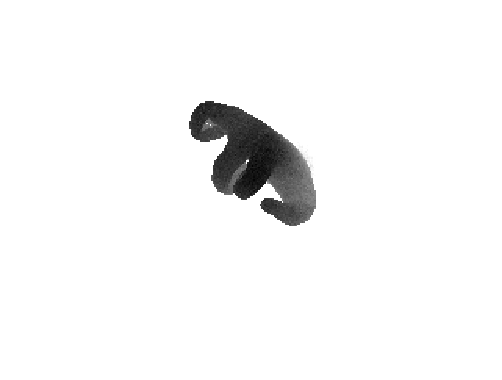}		
	
   \vspace{-6mm}
   
	 \includegraphics[trim=3cm 1cm 2cm 1cm, clip=true,width=0.09\textwidth]{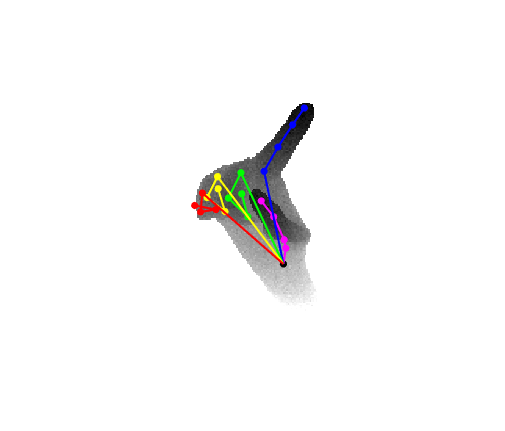}
	 \includegraphics[trim=2.5cm 0.3cm 2.5cm 1.7cm, clip=true,width=0.09\textwidth]{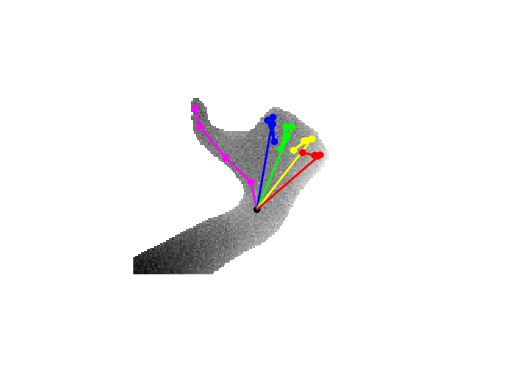}
	 \includegraphics[trim=2.3cm 1.5cm 2.7cm 0.5cm, clip=true,width=0.09\textwidth]{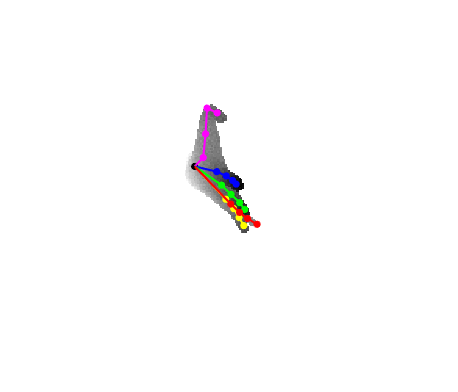}
	 \includegraphics[trim=3cm 1.5cm 3cm 0.5cm, clip=true,width=0.09\textwidth]{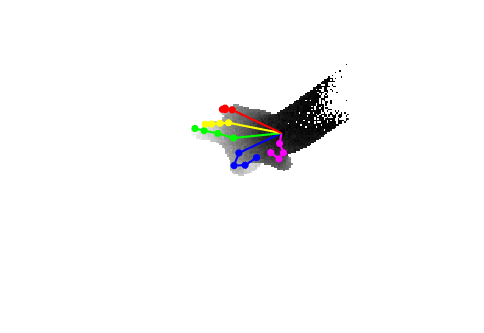}
	 \includegraphics[trim=2.5cm 0.5cm 2.5cm 1.5cm, clip=true,width=0.09\textwidth]{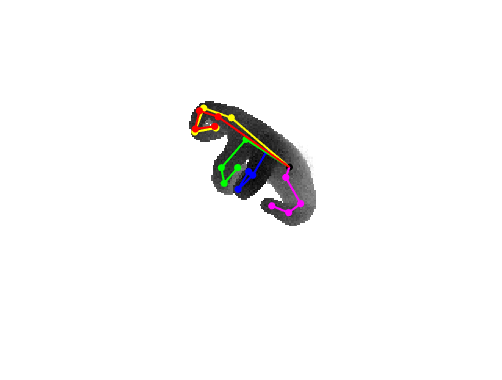}
	
	 \vspace{-5mm}
	
	 \caption{{\bf MSRC benchmark examples}. Synthetic data lacks real hand shape and sensor noise, and tends to have kinematically implausible hand poses. The top row shows some depth images, the bottom row shows the corresponding ground truth annotation.}

     \label{pic:msrcgt}
\end{figure}

\subsection{Cross-benchmark performance}
\label{para:crossbenchmark}

Cross-benchmark evaluation is a challenging and less-studied problem in many fields, like face recognition \cite{parkhi2015deep} and hand pose estimation \cite{supancic2015depth}. Due to the small number of training datasets, existing hand pose estimation systems perform poorly when tested on unseen hand poses. As pointed out in \cite{supancic2015depth}, in existing datasets, ``test poses remarkably resemble the training poses", and they proposed ``a simple nearest-neighbor base line that outperforms most existing systems".

Table~\ref{tab:crossbenchmark} and Figure~\ref{pic:crossbenchmark} show the estimation errors of the CNNs trained on ICVL, NYU, MSRC and {\it BigHand2.2M} when cross-tested. The performance when the CNN is trained on the{\it BigHand2.2M} training set is still high when evaluated on other datasets. On real test datasets (ICVL and NYU), it achieves comparable or even better performance than models trained on the corresponding training set. This confirms that with high annotation accuracy and with sufficient variation in shape, articulation and viewpoint parameters, a CNN  trained on a large-scale dataset is able to generalize to new hand shapes and viewpoints, while the nearest neighbor method showed poor cross-testing performance~\cite{supancic2015depth}.

The MSRC dataset is a synthetic dataset with accurate annotations and evenly distributed viewpoints. When training the CNN on MSRC and testing on all real testing sets, the performance is worse than the CNN trained on NYU, and significantly worse than when trained on {\it BigHand2.2M}. Performance is to that of a CNN trained on ICVL which is only one-sixth in size compared to the MSRC training set. On the other hand, the model trained on  {\it BigHand2.2M} shows consistently high performance across all real datasets, but poorer performance on the MSRC test set
due to the differences between real and synthetic data.
Figure~\ref{pic:msrcgt} shows examples from the MSRC dataset. Synthetically generated images tend to produce kinematically implausible hand poses, which are difficult to assume without applying external force. There are also differences in hand shape, \eg the thumb appears large compared to the rest of the hand.

Increasing the amount of training data improves the performance on cross benchmark evaluation, see Figure~\ref{pic:datasizeeffect}.  In this experiment, we uniformly subsample fractions of $\frac{1}{16}$, $\frac{1}{8}$, $\frac{1}{4}$, $\frac{1}{2}$, and  $1$  from the training and validation data, respectively. When we train CNNs with the increasing portions of {\it BigHand2.2M} and test them on ICVL, NYU, MSRC, and {\it BigHand2.2M}'s test sequences, the performance is fairly improved. These observations support that the larger amount of training data enables CNNs to better generalize to new unseen data. Also, note our dataset is dense such that the random small fractions of the training data still delivers the good accuracies.

\begin{figure*}[t]
\begin{center}
    \includegraphics[width=0.33\textwidth]{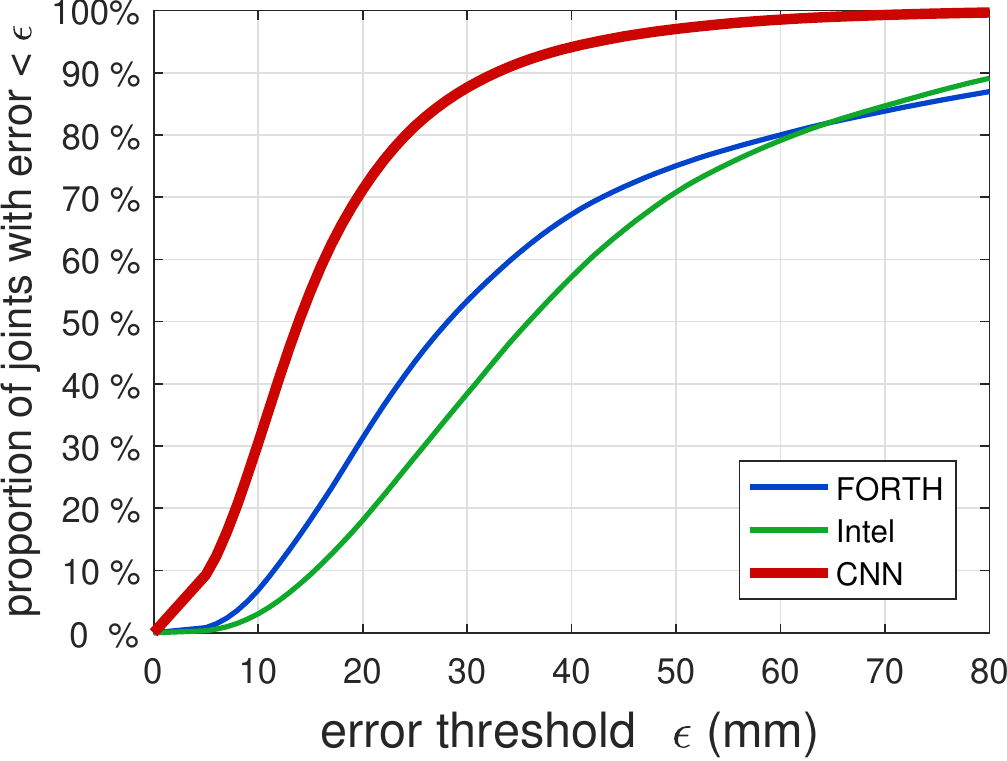}
		\includegraphics[width=0.33\textwidth]{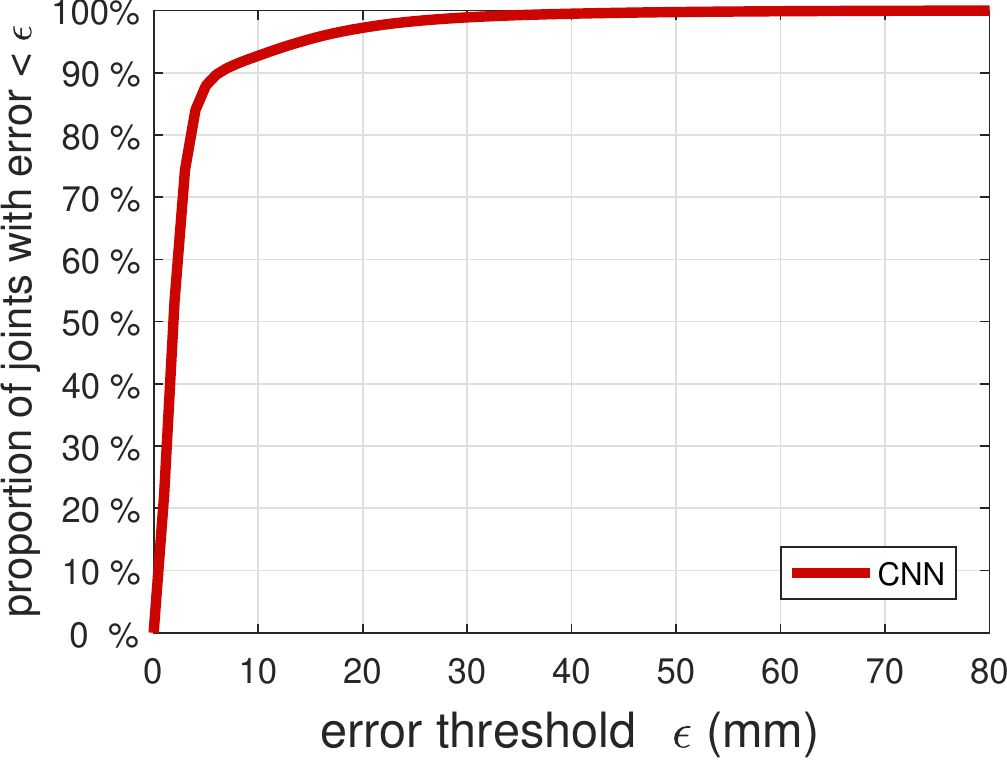}
		\includegraphics[width=0.33\textwidth]{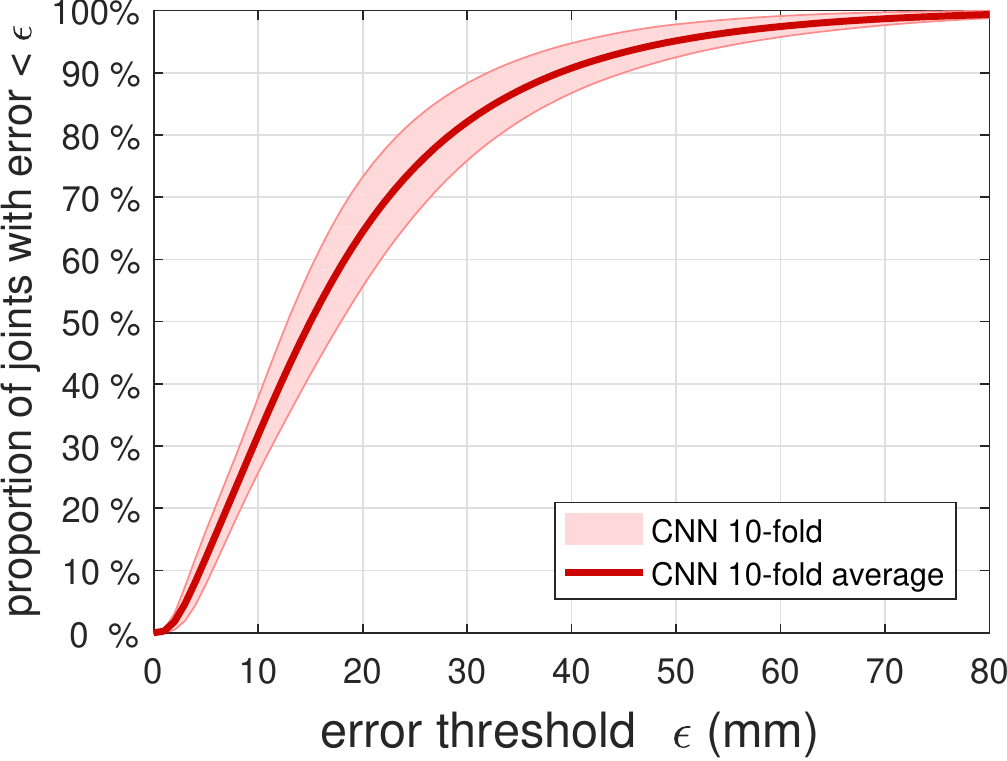}  \hfill \\	
    \caption{{\bf Hand pose estimation performance}. (left) baseline performances on the new subject's 37K frames of hand images. The {\it Holi} CNN significantly outperforms the tracking-based methods FORTH \cite{oikonomidis2011efficient} and Intel \cite{intelSR300}. (middle) the CNN trained on 90\% of the {\it BigHand2.2M} data achieves high accuracy on the remaining 10\% validation images. (right)  10-fold cross-validation result when using the CNN  for egocentric hand pose estimation. We achieved a similar-level accuracy to that of third-view hand pose estimation.}

	\label{pic:baselines}
\end{center}
\end{figure*}

\subsection{State-of-the-art comparison}
\label{para:stateoftheart}

In this section we compare our CNN model trained on {\it BigHand2.2M} with 8 state-of-the-art methods including HSO \cite{tang2015opening}, Sun \textit{et al.} \cite{sun2015cascade}, Latent Regression Forest (LRF) \cite{tang2014latent}, Keskin \textit{et al.} \cite{keskin2012hand},  Melax \textit{et al.} \cite{melax2013dynamics}, DeepPrior \cite{oberweger2015hands}, FeedLoop \cite{oberweger2015training}, and Hier \cite{ye2016spatial}.

When the CNN model trained on {\it BigHand2.2M} is used for testing on NYU, it outperforms two recent methods, DeepPrior \cite{oberweger2015hands} and FeedLoop \cite{oberweger2015training}, and achieves comparable accuracy with Hier~\cite{ye2016spatial}, even though the model has never seen any data from NYU benchmark, demonstrated in the right of Figure~\ref{pic:crossbenchmark}. Since the annotation scheme of NYU is different from ours, we choose a common (still deviated to a certain degree) subset of 11 joint locations for this comparison. We expect better results for consistent annotation schemes.

The ICVL test error curve of the CNN model trained on {\it BigHand2.2M} is shown in Figure~\ref{pic:crossbenchmark}(left). We choose the maximum allowed error \cite{tang2015opening} metric. Although it does not appear as good as that trained on ICVL itself, HSO and Sun et al., it outperforms the other methods. Note that the mean estimation error for our CNN model is already as low as 14mm, which means that a small annotation discrepancy between training and test data will have a large influence on the result. As noted in \cite{oberweger2016efficiently}, the annotation of ICVL is not as accurate as that of NYU. Many frames of our estimation results look plausible, but result in larger estimation errors because of inaccurate annotations, see Figure~\ref{pic:icvlgt} for qualitative comparisons. Another reason is that the hand measurement scheme is different from ours. In our dataset, each subject's hand shape is determined by  manually measuring joint distances. In ICVL, the same synthetic model is used for all subjects and the MCP joints tend to slide towards the fingers rather than remaining on the physical joints. 

\subsection{Baselines on BigHand2.2M}
\label{para:baselines}

Three baselines are evaluated on our 37K-frame testing sequence, the CNN trained on {\it BigHand2.2M}, the Particle Swarm Optimization method (FORTH)~\cite{oikonomidis2011efficient} and the method by Intel~\cite{intelSR300}. The latter two are generative tracking methods. The CNN model outperforms the two generative methods, see the left plot of Figure~\ref{pic:baselines}. As described in the above, we chose a size ratio between training and validation sets of 9:1. Figure~\ref{pic:baselines}(middle) shows the result on the validation set, where 90\% of the joints can be estimated within a 5mm error bound. 

\begin{figure}[t]
\begin{center}

		\includegraphics[trim=1.5cm 0.5cm 1.5cm 0.5cm, clip=true,width=0.09\textwidth]{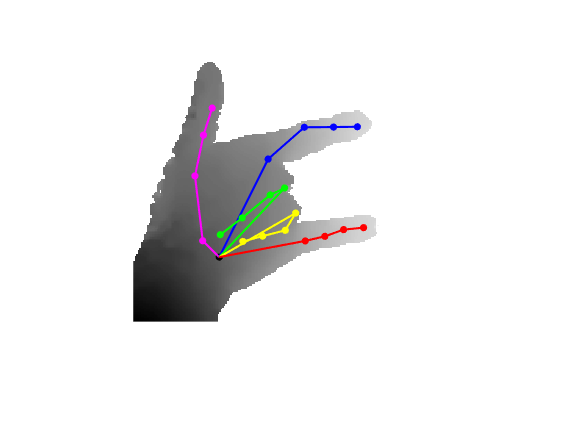}
		\includegraphics[trim=1.5cm 0.5cm 1.5cm 0.5cm, clip=true,width=0.09\textwidth]{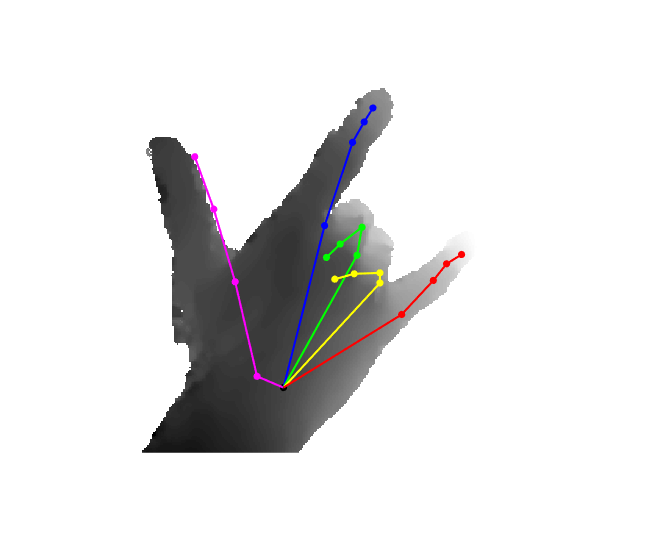}
		\includegraphics[trim=1.5cm 0.5cm 1.5cm 0.5cm, clip=true,width=0.09\textwidth]{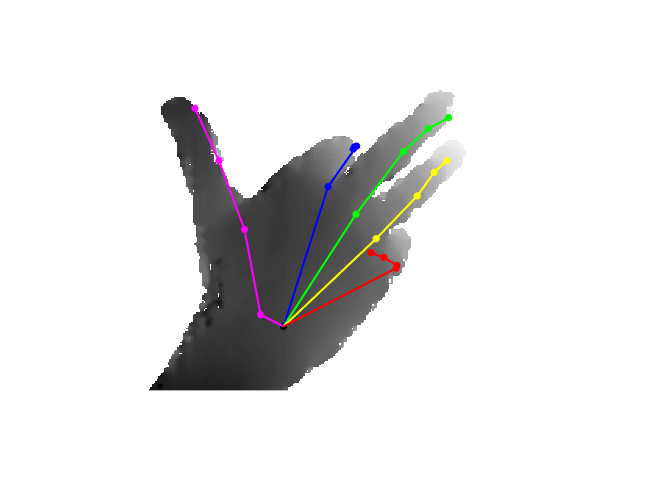}
		\includegraphics[trim=1.5cm 0.5cm 1.5cm 0.5cm, clip=true,width=0.09\textwidth]{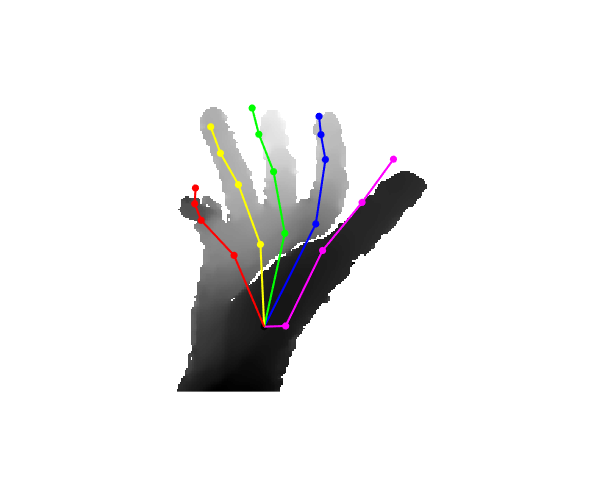}		
		\includegraphics[trim=1.5cm 0.5cm 1.5cm 0.5cm, clip=true,width=0.09\textwidth]{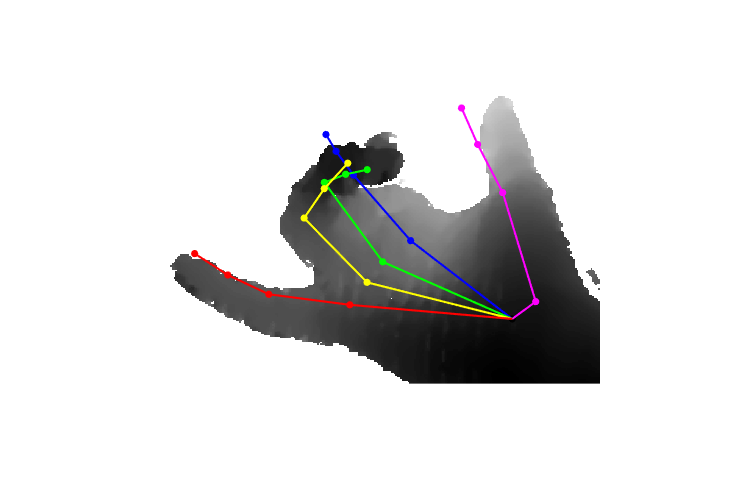}\hfill \\	

		\includegraphics[trim=1.5cm 0.5cm 1.5cm 0.5cm, clip=true,width=0.09\textwidth]{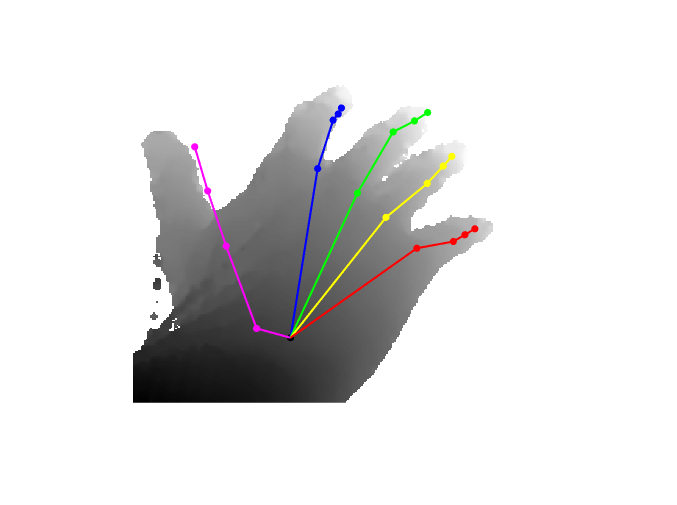}
		\includegraphics[trim=1.5cm 0.5cm 1.5cm 0.5cm, clip=true,width=0.09\textwidth]{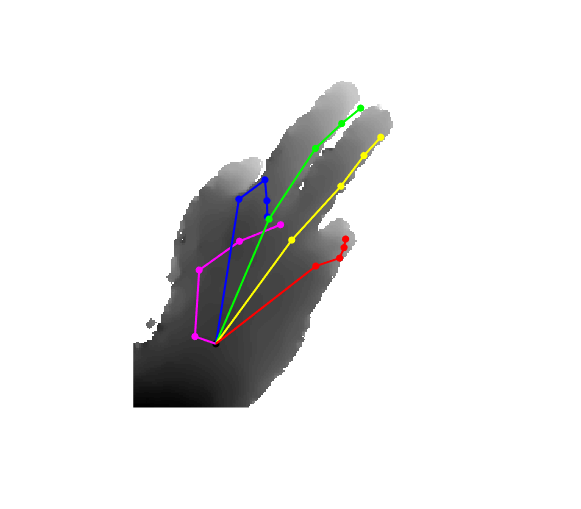}
		\includegraphics[trim=1.5cm 0.5cm 1.5cm 0.5cm, clip=true,width=0.09\textwidth]{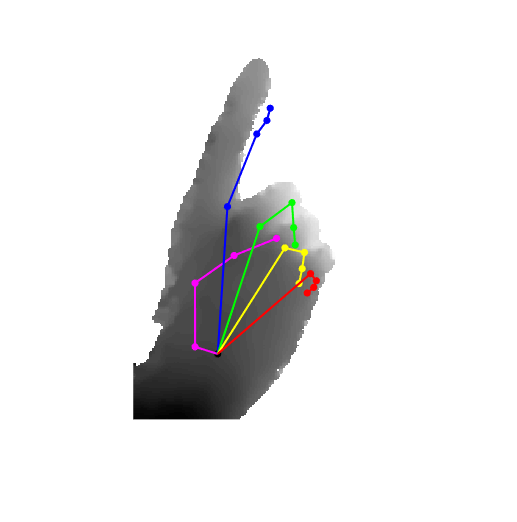}
		\includegraphics[trim=1.5cm 0.5cm 1.5cm 0.5cm, clip=true,width=0.09\textwidth]{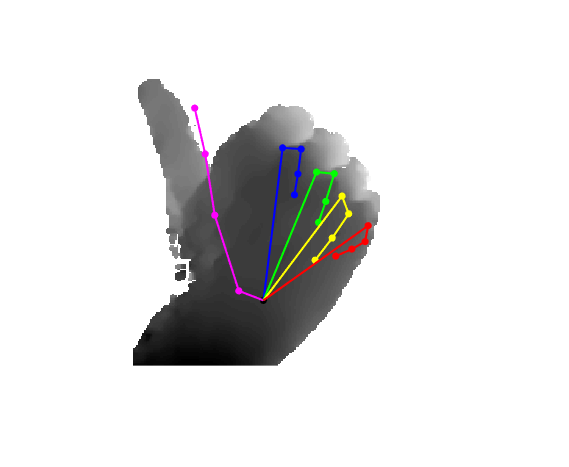}		
		\includegraphics[trim=1.5cm 0.5cm 1.5cm 0.5cm, clip=true,width=0.09\textwidth]{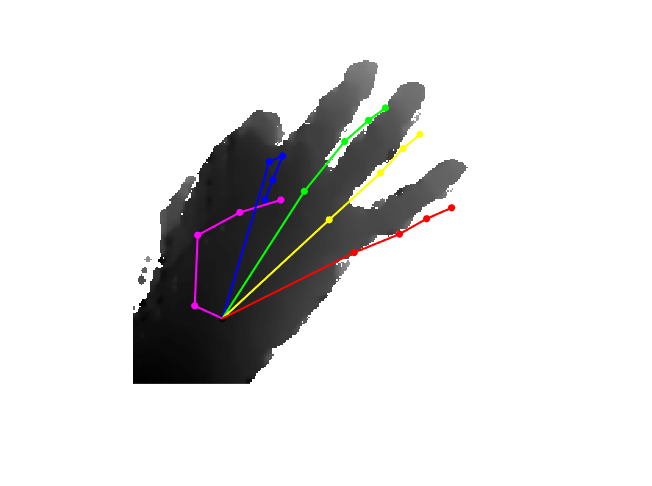}\hfill \\

		\includegraphics[trim=1.5cm 0.5cm 1.5cm 0.5cm, clip=true,width=0.09\textwidth]{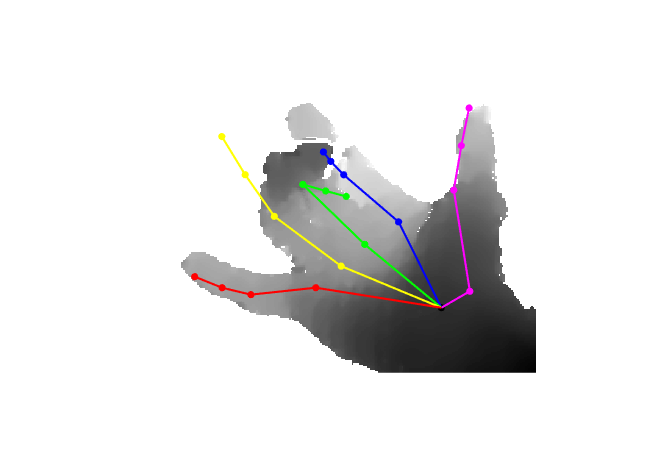}
		\includegraphics[trim=1.5cm 0.5cm 1.5cm 0.5cm, clip=true,width=0.09\textwidth]{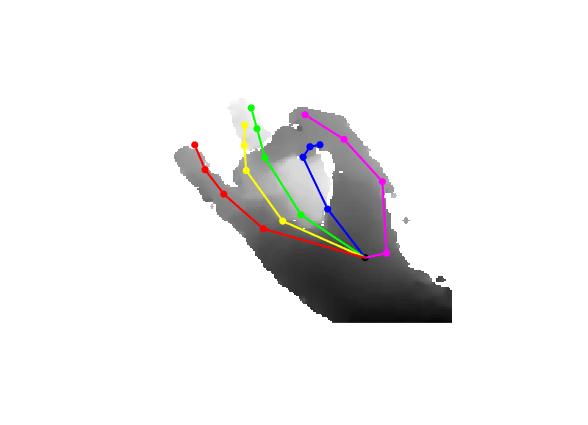}
    \includegraphics[trim=1.5cm 0.5cm 1.5cm 0.5cm, clip=true,width=0.09\textwidth]{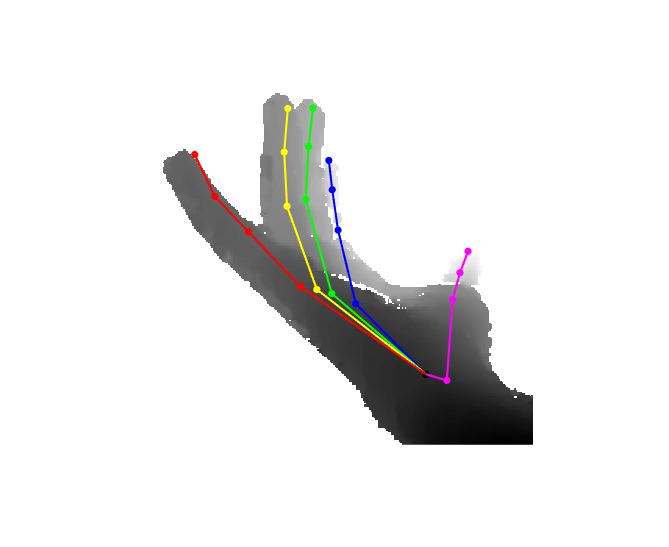}		
		\includegraphics[trim=1.5cm 0.5cm 1.5cm 0.5cm, clip=true,width=0.09\textwidth]{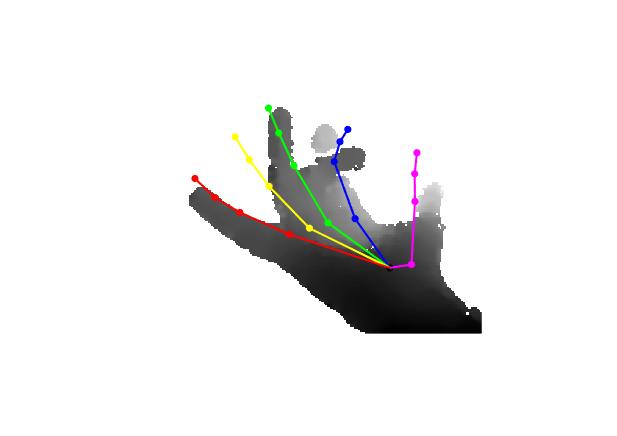}
		\includegraphics[trim=1.5cm 0.5cm 1.5cm 0.5cm, clip=true,width=0.09\textwidth]{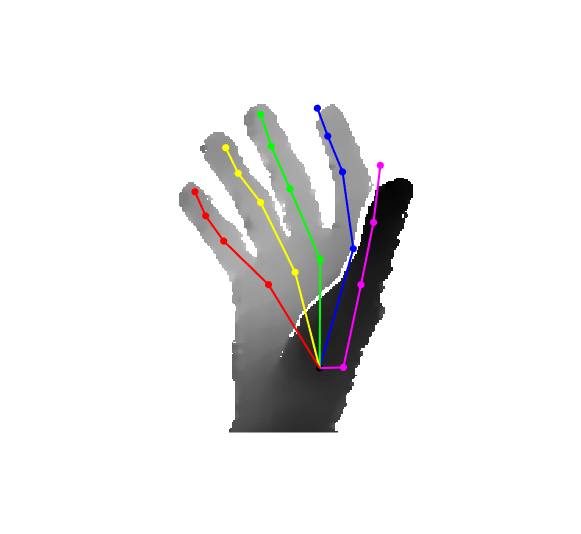}
		\hfill \\	

    \caption{{\bf Qualitative results on the egocentric-view dataset}. A CNN trained on {\it BigHand2.2M} achieves state-of-the-art performance in the egocentric-view pose estimation task.}

	\label{pic:egoresultexamples}
\end{center}
\end{figure}

\subsection{Egocentric dataset}
\label{para:egocentric}

The lack of a large-scale annotated dataset has been a limiting factor for egocentric hand pose estimation. Existing egocentric benchmarks \cite{rogez2015first,oberweger2016efficiently} are small, see Table~\ref{tab:egobenchmarkcomp}. Rogez \etal \cite{rogez2015first} provide 400 frames and Oberwerger \etal \cite{oberweger2016efficiently} provide 2,166 annotated frames. The {\it BigHand2.2M} egocentric subset contains 290K annotated frames of ten subjects (29K frames each). This dataset enabled us to train a CNN model resulting in performance competitive with that of third view hand pose estimation. In the experiment we train the CNN on nine subjects and test it on the remaining one. This process is done with 10-fold cross validation. We report mean and standard deviation of the ten folds, see Figure~\ref{pic:baselines} (right). Figure~\ref{pic:egoresultexamples} shows qualitative results.

\section{Discussion and conclusion}
\label{para:conc}

Hand pose estimation has attracted a lot of attention and some high-quality systems have been demonstrated, but the development in datasets still lagged behind the algorithm advancement. To close this gap we captured a million-scale benchmark dataset of real hand depth images. For automatic annotation we proposed using a magnetic tracking system with six magnetic 6D sensors and inverse kinematics. To build a thorough yet concise benchmark, we systematically designed a hand movement protocol to capture the natural hand poses. The {\it BigHand2.2M} dataset includes approximately 290K frames captured from an egocentric view to facilitate the advancement in the area of egocentric hand pose estimation. Current state-of-the-art methods were evaluated using the new benchmark, and we demonstrated significant improvements in cross-benchmark evaluations. It is our aim that the dataset will help to further advance the research field, allowing the exploration of new approaches.

{\small
\bibliographystyle{ieee}
\bibliography{egbib}

\begin{thebibliography}{10}\itemsep=-1pt

\bibitem{choi2015collaborative}
C.~Choi, A.~Sinha, J.~Hee~Choi, S.~Jang, and K.~Ramani.
\newblock A collaborative filtering approach to real-time hand pose estimation.
\newblock In {\em ICCV, 2015}.

\bibitem{ge2016robust}
L.~Ge, H.~Liang, J.~Yuan, and D.~Thalmann.
\newblock Robust 3d hand pose estimation in single depth images: from
  single-view cnn to multi-view cnns.
\newblock In {\em CVPR, 2016}.

\bibitem{intelSR300}
{Intel SR300}.
\newblock
  \url{https://click.intel.com/intelrealsense-developer-kit-featuring-sr300.html}.

\bibitem{jang20153d}
Y.~Jang, S.-T. Noh, H.~J. Chang, T.-K. Kim, and W.~Woo.
\newblock 3d finger cape: Clicking action and position estimation under
  self-occlusions in egocentric viewpoint.
\newblock {\em VR, 2015}.

\bibitem{keskin2012hand}
C.~Keskin, F.~K{\i}ra{\c{c}}, Y.~E. Kara, and L.~Akarun.
\newblock Hand pose estimation and hand shape classification using
  multi-layered randomized decision forests.
\newblock In {\em ECCV, 2012}.

\bibitem{khamis2015learning}
S.~Khamis, J.~Taylor, J.~Shotton, C.~Keskin, S.~Izadi, and A.~Fitzgibbon.
\newblock Learning an efficient model of hand shape variation from depth
  images.
\newblock In {\em CVPR, 2015}.

\bibitem{li20153d}
P.~Li, H.~Ling, X.~Li, and C.~Liao.
\newblock 3d hand pose estimation using randomized decision forest with
  segmentation index points.
\newblock In {\em ICCV, 2015}.

\bibitem{liang2014parsing}
H.~Liang, J.~Yuan, and D.~Thalmann.
\newblock Parsing the hand in depth images.
\newblock {\em TMM, 2014}.

\bibitem{melax2013dynamics}
S.~Melax, L.~Keselman, and S.~Orsten.
\newblock Dynamics based {3D} skeletal hand tracking.
\newblock In {\em i3D, 2013}.

\bibitem{trakSTAR}
{NDI trakSTAR}.
\newblock \url{https://www.ascension-tech.com/products/trakstar-2-drivebay-2/}.

\bibitem{neverova2015hand}
N.~Neverova, C.~Wolf, G.~W. Taylor, and F.~Nebout.
\newblock Hand segmentation with structured convolutional learning.
\newblock In {\em ACCV, 2014}.

\bibitem{oberweger2016efficiently}
M.~Oberweger, G.~Riegler, P.~Wohlhart, and V.~Lepetit.
\newblock Efficiently creating {3D} training data for fine hand pose
  estimation.
\newblock In {\em CVPR, 2016}.

\bibitem{oberweger2015hands}
M.~Oberweger, P.~Wohlhart, and V.~Lepetit.
\newblock Hands deep in deep learning for hand pose estimation.
\newblock In {\em CVWW, 2015}.

\bibitem{oberweger2015training}
M.~Oberweger, P.~Wohlhart, and V.~Lepetit.
\newblock Training a feedback loop for hand pose estimation.
\newblock In {\em ICCV, 2015}.

\bibitem{oikonomidis2011efficient}
I.~Oikonomidis, N.~Kyriazis, and A.~A. Argyros.
\newblock Efficient model-based {3D} tracking of hand articulations using
  kinect.
\newblock In {\em BMVC}, 2011.

\bibitem{parkhi2015deep}
O.~M. Parkhi, A.~Vedaldi, and A.~Zisserman.
\newblock Deep face recognition.
\newblock In {\em BMVC, 2015}.

\bibitem{pons2011outdoor}
G.~Pons-Moll, A.~Baak, J.~Gall, L.~Leal-Taixe, M.~Mueller, H.-P. Seidel, and
  B.~Rosenhahn.
\newblock Outdoor human motion capture using inverse kinematics and von
  mises-fisher sampling.
\newblock In {\em ICCV, 2011}.

\bibitem{qian2014realtime}
C.~Qian, X.~Sun, Y.~Wei, X.~Tang, and J.~Sun.
\newblock Realtime and robust hand tracking from depth.
\newblock In {\em CVPR, 2014}.

\bibitem{riegler2015framework}
G.~Riegler, D.~Ferstl, M.~R{\"u}ther, and H.~Bischof.
\newblock A framework for articulated hand pose estimation and evaluation.
\newblock In {\em SCIA, 2015}.

\bibitem{rogez2015first}
G.~Rogez, J.~S. Supancic, and D.~Ramanan.
\newblock First-person pose recognition using egocentric workspaces.
\newblock In {\em CVPR, 2015}.

\bibitem{rogez2015understanding}
G.~Rogez, J.~S. Supancic, and D.~Ramanan.
\newblock Understanding everyday hands in action from {RGB-D} images.
\newblock In {\em ICCV, 2015}.

\bibitem{schaffelhofer2012new}
S.~Schaffelhofer and H.~Scherberger.
\newblock A new method of accurate hand- and arm-tracking for small primates.
\newblock {\em Journal of Neural Engineering}, 9(2), 2012.

\bibitem{ShapeHand}
ShapeHand.
\newblock 2009.
\newblock \url{http://www.shapehand.com/shapehand.html}.

\bibitem{sharp2015accurate}
T.~Sharp, C.~Keskin, D.~Robertson, J.~Taylor, J.~Shotton, D.~Kim, C.~Rhemann,
  I.~Leichter, A.~Vinnikov, Y.~Wei, D.~Freedman, P.~Kohli, E.~Krupka,
  A.~Fitzgibbon, and S.~Izadi.
\newblock Accurate, robust, and flexible real-time hand tracking.
\newblock In {\em CHI, 2015}.

\bibitem{handtracker_iccv2013}
S.~Sridhar, A.~Oulasvirta, and C.~Theobalt.
\newblock Interactive markerless articulated hand motion tracking using rgb and
  depth data.
\newblock In {\em ICCV, 2013}.

\bibitem{sun2015cascade}
X.~Sun, Y.~Wei, S.~Liang, X.~Tang, and J.~Sun.
\newblock Cascaded hand pose regression.
\newblock In {\em CVPR, 2015}.

\bibitem{supancic2015depth}
J.~S. Supancic, G.~Rogez, Y.~Yang, J.~Shotton, and D.~Ramanan.
\newblock Depth-based hand pose estimation: methods, data, and challenges.
\newblock {\em ICCV, 2015}.

\bibitem{tang2014latent}
D.~Tang, H.~J. Chang, A.~Tejani, and T.-K. Kim.
\newblock Latent regression forest: Structured estimation of {3D} articulated
  hand posture.
\newblock In {\em CVPR}, 2014.

\bibitem{tang2015opening}
D.~Tang, J.~Taylor, P.~Kohli, C.~Keskin, T.-K. Kim, and J.~Shotton.
\newblock Opening the black box: Hierarchical sampling optimization for
  estimating human hand pose.
\newblock In {\em ICCV, 2015}.

\bibitem{tang2013real}
D.~Tang, T.-H. Yu, and T.-K. Kim.
\newblock Real-time articulated hand pose estimation using semi-supervised
  transductive regression forests.
\newblock In {\em ICCV}, 2013.

\bibitem{tompson2014real}
J.~Tompson, M.~Stein, Y.~Lecun, and K.~Perlin.
\newblock Real-time continuous pose recovery of human hands using convolutional
  networks.
\newblock In {\em TOG, 2014}.

\bibitem{von2016human}
T.~von Marcard, G.~Pons-Moll, and B.~Rosenhahn.
\newblock Human pose estimation from video and imus.
\newblock {\em TPAMI, 2016}.

\bibitem{wan2016hand}
C.~Wan, A.~Yao, and L.~Van~Gool.
\newblock Hand pose estimation from local surface normals.
\newblock In {\em ECCV, 2016}.

\bibitem{wetzler2015rule}
A.~Wetzler, R.~Slossberg, and R.~Kimmel.
\newblock Rule of thumb: Deep derotation for improved fingertip detection.
\newblock In {\em BMVC, 2015}.

\bibitem{wu2001capturing}
Y.~Wu, J.~Y. Lin, and T.~S. Huang.
\newblock Capturing natural hand articulation.
\newblock In {\em ICCV, 2001}.

\bibitem{XuEtAl_IJCV15}
C.~Xu, N.~Ashwin, X.~Zhang, and L.~Cheng.
\newblock Estimate hand poses efficiently from single depth images.
\newblock {\em IJCV}, 2015.

\bibitem{xu2013efficient}
C.~Xu and L.~Cheng.
\newblock Efficient hand pose estimation from a single depth image.
\newblock In {\em ICCV}, 2013.

\bibitem{ye2016spatial}
Q.~Ye, S.~Yuan, and T.-K. Kim.
\newblock Spatial attention deep net with partial {PSO} for hierarchical hybrid
  hand pose estimation.
\newblock In {\em ECCV, 2016}.

\bibitem{zheng2013aspnp}
Y.~Zheng, S.~Sugimoto, and M.~Okutomi.
\newblock {ASPnP}: An accurate and scalable solution to the perspective-n-point
  problem.
\newblock {\em IEICE TIS, 2013}.

\bibitem{zhou2016model}
X.~Zhou, Q.~Wan, W.~Zhang, X.~Xue, and Y.~Wei.
\newblock Model-based deep hand pose estimation.
\newblock In {\em IJCAI, 2016}.

\end{thebibliography}
}

\end{document}